\documentclass{jamia}

\usepackage{amsmath,amssymb,amsfonts}%
\usepackage{booktabs}%
\usepackage{colortbl}
\usepackage{enumitem}
\usepackage{float}
\usepackage{graphicx}%
\usepackage[colorlinks=true]{hyperref}
\usepackage{listings}%
\usepackage{makecell}
\usepackage{mdframed}
\usepackage[version=4]{mhchem}
\usepackage{minted}
\usepackage[numbers,comma,sort&compress]{natbib} 
\usepackage{subcaption}
\usepackage{tabularray}
\usepackage{tabularx}
\usepackage[most]{tcolorbox}
\usepackage{xcolor}%
\usepackage{xspace}

\definecolor{graybg}{HTML}{F5F5F5}
\newcommand\boxedname{Box\xspace} 
\newcounter{prompt}
\newenvironment{prompt}[1][]{
    \refstepcounter{prompt}
    \definecolor{graybg}{HTML}{f1f1f1}
    \begin{mdframed}[
        innertopmargin=2pt, 
        innerbottommargin=2pt,
        innerleftmargin=2pt, 
        innerrightmargin=2pt,
        frametitle={\textbf{\boxedname \theprompt:} #1},
        frametitlefont=\scriptsize,
        frametitlerule=true,
        backgroundcolor=graybg]%
    \setlength{\parindent}{0pt}%
    \setlength{\parskip}{.5em}%
    \setlist[itemize]{nosep}
    \setlist[enumerate]{nosep}

    \scriptsize\ttfamily\hyphenchar\font=`\-\spaceskip=.5em plus .5em\xspaceskip=.5em%
}
{%
    \par%
    \end{mdframed}%
}

\lstdefinelanguage{prompt}{
    breaklines=true,
    breakindent=0pt,
    basicstyle=\small\ttfamily,
    columns=fullflexible,
    upquote=true,
}

\newlist{promptenum}{enumerate}{1}
\setlist[promptenum]{label=\textbf{\alph*.}, nosep, leftmargin=1.5em, labelwidth=2em, labelsep=0em}

\newlist{promptitem}{itemize}{1}
\setlist[promptitem]{label=\labelitemi, nosep, leftmargin=1.5em, labelwidth=2em, labelsep=.75em}

\newcommand{\Ourmodel}{LLMSurvival\xspace}

\title{Towards end-to-end LLM-based censoring-aware survival analysis}

\author[1,$\dagger$]{Yishu Wei, Ph.D.}
\author[1,$\dagger$]{Hexin Dong, Ph.D.}
\author[1]{Yi Lin, Ph.D.}
\author[1]{Jiahe Qian, M.S.}
\author[2]{Yi Liu, M.D.}
\author[1,*]{Yifan Peng, Ph.D.}
\affil[1]{Department of Population Health Science, Weill Cornell Medicine, New York, 10065, US}
\affil[2]{Department of Medicine, Weill Cornell Medicine, New York, 10065, US}

\affil[$\dagger$]{These authors contributed equally to this work.}
\affil[*]{Corresponding author(s). Email(s): \url{yip4002@med.cornell.edu}}

\begin{document}
\maketitle

\begin{abstract}
\textbf{Objective:} Survival analysis is central to medical prediction, yet large language models (LLMs) are rarely used as end-to-end survival models because censoring prevents straightforward supervised fine-tuning. Here we present LLMSurvival, a framework that enables censoring-aware survival analysis with unmodified LLMs operating directly on tabular clinical data.

\textbf{Materials and Methods:} LLMSurvival reformulates time-to-event prediction as pairwise ranking among comparable subjects, and derives test-time risk by aggregating comparisons against anchor individuals from the training cohort.

\textbf{Results:} Across two clinical tasks (ICU mortality prediction in MIMIC-IV and fragility fracture prediction in a NewYork-Presbyterian/Weill Cornell Medicine cohort), LLMSurvival improves overall concordance over Cox proportional hazards modeling by 3.1\% for ICU mortality and 0.5\% for fracture risk, and by 2.1\% on average for ICU mortality and 2.8\% for fracture risk over three established deep learning survival models. 

\textbf{Discussion:} The results show that survival modeling with censoring can be made compatible with LLM fine-tuning through comparison-based reformulation. The framework demonstrates high portability and superior performance over expert-curated scores like SAPS-II and FRAX scores across diverse clinical contexts. Furthermore, the framework supports local deployment, as compact, publicly available base models provide sufficient performance.

\textbf{Conclusion:} The LLMSurvival framework serves as a proof of concept for an integrated, censoring-conscious approach to survival analysis via LLMs.
\end{abstract}

\section{Background and Significance}

Survival analysis is a central problem in statistics and machine learning, particularly within the medical domain. Its objective is to model not only whether an event will occur but also the time until its occurrence \cite{chung1991survival}. Although extensive research has explored the application of LLMs to survival analysis, the majority of these studies utilize LLMs merely as feature extractors rather than direct predictive models, or require substantial architectural modifications  \cite{jeanselme2024review}. 
\citet{shahid2026leveraging} used an LLM as a feature extractor to obtain variables from clinical notes. 
\citet{vaidyanathan2025survival} simplified the survival analysis problem into a five-year classification task and did not address censoring. 
\citet{wen2024llm} also used an LLM to extract features for a survival model. \citet{steinberg2023motor} used a piecewise-exponential artificial neural network to predict the hazard rate. 
Similarly, \citet{esteban2024predictive} added dedicated neural networks on top of an LLM and experimented with ordinal regression loss and discordant pair loss. When neural networks are added on top of an LLM, the LLMs still serve primarily as feature extractors. For a summary of the previous literature, please refer to eTable \ref{tab:survival_analysis_literature} in Supplement 1.

Large language models (LLMs) have been directly employed as end-to-end models for classical statistical learning tasks without architectural modifications. 
For instance, \citet{dinh2022lift} formatted a regression task on tabular data into natural language prompts. They used the template ``When we have $x_1=r.x1, x2=r.x2, ..., xp=r.xp$, what should be y? \#\#\# $y=r.y$ @@@" to format data and fine-tune LLMs based on that. Similarly, \citet{hegselmann2023tabllm} fine-tuned LLMs for classification by framing problems as binary ``Yes/No" questions, achieving performance comparable to traditional statistical methods. However, such strategies cannot be readily extended to survival analysis due to the challenges of censoring. Censoring occurs when the exact time of an event is unknown, typically because a subject withdraws from a study or the observation period concludes before the event occurs. Discarding censored subjects introduces significant bias \cite{wang2019machine}. Conversely, including them is difficult because they lack explicit labels. This lack of ground truth poses a fundamental hurdle for LLM training, particularly in the widely adopted supervised fine-tuning and reinforcement learning frameworks.

In this study, we aim to bridge a gap in the literature and address the challenge that LLMs cannot be directly optimized for survival analysis due to the lack of explicit labels caused by censoring. We present \Ourmodel, a multi-stage framework designed to apply LLMs to survival analysis directly on tabular data. Rather than training the LLM to directly predict survival times, we optimize the model for pairwise comparison. During inference, each subject is compared against a set of pre-selected anchor subjects. The risk score is then estimated based on the proportion of comparisons in which the test subject is predicted to experience the event earlier. By reformulating survival analysis as a ranking problem, we naturally circumvent the challenges of censoring and can use all comparable pairs. Furthermore, this ranking approach leverages the inherent strengths of LLMs, as comparing two subject profiles described in natural language is a task well-suited to their contextual reasoning.

To better understand how censoring causes data loss in classification-based survival modeling and how a ranking formulation can mitigate this problem, consider a concrete example with three subjects: A (event at 5 years), B (event at 10 years), and C(censored at 8 years). If the prediction task is framed as a binary classification problem at a fixed time horizon at year 9 or later, Subject C must be excluded since its status after year 8 is unknown and thus cannot be assigned an explicit label. 

However, if the task is framed as a ranking-based problem, we can construct comparable pairs from the observed survival data. In this example, two valid pairs can be formed. First, Subject A is observed to experience the event before Subject B. Second, Subject A is also observed to experience the event before Subject C is censored, as C is known to remain event-free at year 5. However, Subjects B and C are not comparable because C is censored at year 8, leaving its true status at year 10 (when B's event occurs) unknown. By utilizing this pairwise approach, Subject C is preserved in the dataset and contributes partially informative comparisons, preventing data loss and reducing bias.

Treating survival analysis as a ranking problem has been widely explored in prior machine learning and deep learning work \cite{Jing2019-nj, Van-Belle2011-ab,luck2018learning,jing2019deep}. However, there is limited research extending these explorations to large language models. LLMs have independently shown strong performance on ranking tasks in natural language processing \cite{Qin2024-eo, chao2024make}. Our work bridges these two lines of research. Pairwise ranking effectively addresses censoring while leveraging LLMs' inherent strengths, positioning this approach at the ideal intersection of survival analysis and natural language processing.

We evaluate \Ourmodel in two diverse clinically important applications: mortality prediction in the Intensive Care Unit (ICU) and fracture risk assessment. Across both settings, \Ourmodel achieves performance that is better than traditional and deep-learning-based survival methods (e.g., Cox proportional hazards model, Cox-nnet \cite{ching2018cox}, Deephit \cite{lee2018deephit} and DeepSurv \cite{katzman2018deepsurv}). Furthermore, the model outperforms established clinical risk scores, including SAPS II for ICU mortality and FRAX for fracture risk. \Ourmodel also exhibits consistent performance trends across follow-up intervals. 

The main contribution of this study is to introduce a framework that can directly train an LLM as the modeler for survival analysis. It expands its role from mere feature extractors or bottom layers of the overall architecture to a comprehensive, end-to-end modeling pipeline. It serves as a proof of concept, demonstrating that large language models can expand their capabilities to solve more complex statistical problems on their own. In addition, it demonstrates that reformulating a complex problem (such as survival analysis) into a task better suited to LLMs (e.g., pairwise comparison) enables them to tackle challenges they could not otherwise solve independently.

\paragraph*{Statement of Significance}
\begin{itemize}[nosep]
    \item \textbf{Problem}: Survival analysis is essential for clinical prognosis, but censoring prevents straightforward supervised fine-tuning of large language models as end-to-end survival models.
    \item \textbf{What is Already Known}: LLMs have been used in biomedical prediction mainly as feature extractors or through simplified classification tasks; conventional and deep survival models address censoring but require survival-specific architectures or assumptions.
    \item \textbf{What This Paper Adds}: We introduce LLMSurvival, which reformulates censored survival prediction as pairwise ranking over comparable patients and estimates risk by aggregating comparisons against training anchors. Across ICU mortality and fragility fracture prediction, it improves discrimination over Cox models and established clinical scores.
    \item \textbf{Who Would Benefit from the New Knowledge in This Paper}: Researchers and clinicians developing deployable clinical risk models for censored time-to-event outcomes.
\end{itemize}

\section{Materials and Methods}
\label{sec:methods}

\subsection{Ethics statement}

The retrospective analysis of internal electronic medical records was reviewed and approved by the Weill Cornell Medicine Institutional Review Board. All data were anonymized before computational analysis. Because the study used retrospective de-identified data and involved no direct patient contact, informed consent was waived.

\subsection{Study design}
\label{sec:dataset}

We evaluated \Ourmodel on two complementary cohorts representing acute and chronic clinical prediction tasks: ICU mortality prediction using MIMIC-ICU (Table \ref{stab:mimic}) and incident fragility fracture prediction using an NYP/WCM cohort (Table \ref{tab:fracture}). For both cohorts, data were split into training, validation, and test sets using a ratio of 7:1:2.

\paragraph{MIMIC-ICU cohort.} 

We used MIMIC-IV (Medical Information Mart for Intensive Care IV) \cite{Johnson2023-sr}, a de-identified electronic health record database of adult ICU admissions at Beth Israel Deaconess Medical Center between 2008 and 2019.
For consistency with previous work, we adopted the cohort definition of \citet{Lin2021-ye}, yielding 9,928 ICU patients. Of these, 2,213 (22\%) deceased during their ICU stay (Table \ref{stab:mimic}). 

We used SAPS-II-based structured variables derived from the first ICU day, including age, physiologic measures, laboratory abnormalities, chronic disease burden, and admission type \cite{leGall1993new}. SAPS-II scores were computed according to the original definition using publicly available code\footnote{\url{https://github.com/MIT-LCP/mimic-code/blob/main/mimic-iii/concepts/severityscores/sapsii.sql}}. 

{\definecolor{stone}{HTML}{BFBA9C}
\definecolor{grey}{HTML}{F6F3EB}
\footnotesize
\begin{longtblr}[
  caption = {Participant demographic and clinical characteristics on the MIMIC-ICU cohort.},
  label = {stab:mimic},
]{
  colspec = {lXlll},
  rowhead = 2,
  hline{1,3,Z} = {0.08em, black},
}

&& \SetCell[c=3]{l}{\textbf{Participants, No. (\%)}} \\
\cline{3-5}
\SetCell[c=2]{l}\textbf{Characteristic}
& & \textbf{All (N=9928)}
& \textbf{Discharge (n=7715)}
& \textbf{Mortality (n=2213)} \\
%
\SetCell[c=2]{l}Age, mean (SD), y
&& 65.13 (17.99)
& 63.32 (18.38)
& 71.43 (14.93) \\
\hline[stone]
\SetCell[c=2]{l}Gender, male/female, \%
&& 5416 (55) / 4512 (45)
& 4221 (55) / 3494 (45)
& 1195 (54) / 1018 (46) \\
\hline[stone]
\SetCell[c=2]{l}SAPS-II, mean (SD)
&& 37.25 (15.44)
& 33.51 (13.01)
& 50.29 (16.15) \\
\hline[stone]
\SetCell[c=5]{l}\textbf{SAPS-II physiological measurements} \\
\SetCell[c=2]{l}Age, year \\
\cline[stone]{2-5}
& $<$ 40
& 1036 (10.44)
& 965 (12.51)
& 71 (3.21) \\
\cline[stone]{2-5}
& 40-59
& 2390 (24.07)
& 1999 (25.91)
& 391 (17.67) \\
\cline[stone]{2-5}
& 60-69
& 2127 (21.42)
& 1669 (21.63)
& 458 (20.70) \\
\cline[stone]{2-5}
& 70-74
& 969 (9.76)
& 707 (9.16)
& 262 (11.84) \\
\cline[stone]{2-5}
& 75-79
& 957 (9.64)
& 709 (9.19)
& 248 (11.21) \\
\cline[stone]{2-5}
& $\ge$ 80
& 2449 (24.67)
& 1666 (21.59)
& 783 (35.38) \\

\hline[stone]
\SetCell[c=2]{l}Heart Rate \\
\cline[stone]{2-5}
& $<$ 40
& 187 (1.88)
& 80 (1.04)
& 107 (4.84) \\
\cline[stone]{2-5}
& 40-69
& 3884 (39.12)
& 3262 (42.28)
& 622 (28.11) \\
\cline[stone]{2-5}
& 70-119
& 3406 (34.31)
& 2678 (34.71)
& 728 (32.90) \\
\cline[stone]{2-5}
& 120-159
& 2314 (23.31)
& 1623 (21.04)
& 691 (31.22) \\
\cline[stone]{2-5}
& $\ge$ 160
& 137 (1.38)
& 72 (0.93)
& 65 (2.94) \\

\hline[stone]
\SetCell[c=2]{l}Systolic BP, mmHg \\
\cline[stone]{2-5}
& $<$ 70
& 912 (9.19)
& 425 (5.51)
& 487 (22.01) \\
\cline[stone]{2-5}
& 70-99
& 6138 (61.83)
& 4821 (62.49)
& 1317 (59.51) \\
\cline[stone]{2-5}
& 100-199
& 2764 (27.84)
& 2378 (30.82)
& 386 (17.44) \\
\cline[stone]{2-5}
& $\ge$ 200
& 114 (1.15)
& 91 (1.18)
& 23 (1.04) \\

\hline[stone]
\SetCell[c=2]{l}Temperature $\leq$ 39 \textdegree C \\
\cline[stone]{2-5}
& No
& 9449 (95.18)
& 7374 (95.58)
& 2075 (93.76) \\
\cline[stone]{2-5}
& Yes
& 479 (4.82)
& 341 (4.42)
& 138 (6.24) \\

\hline[stone]
\SetCell[c=2]{l}\ce{PaO2 / FiO2}, mmHg \\
\cline[stone]{2-5}
& $<$ 100
& 562 (5.66)
& 268 (3.47)
& 294 (13.29) \\
\cline[stone]{2-5}
& 100-199
& 892 (8.98)
& 584 (7.57)
& 308 (13.92) \\
\cline[stone]{2-5}
& $\ge$ 200
& 1439 (14.49)
& 1054 (13.66)
& 385 (17.40) \\
\cline[stone]{2-5}
& No ventilation
& 7035 (70.86)
& 5809 (75.29)
& 1226 (55.40) \\

\hline[stone]
\SetCell[c=2]{l}Blood urea nitrogen, mg/dL \\
\cline[stone]{2-5}
& $<$ 28
& 6471 (65.18)
& 5422 (70.28)
& 1049 (47.40) \\
\cline[stone]{2-5}
& 28-93
& 3090 (31.12)
& 2068 (26.80)
& 1022 (46.18) \\
\cline[stone]{2-5}
& $\ge$ 84
& 367 (3.70)
& 225 (2.92)
& 142 (6.42) \\

\hline[stone]
\SetCell[c=2]{l}Urine output, mL/day \\
\cline[stone]{2-5}
& $<500$
& 1246 (12.55)
& 591 (7.66)
& 655 (29.60) \\
\cline[stone]{2-5}
& 500-999
& 1738 (17.51)
& 1299 (16.84)
& 439 (19.84) \\
\cline[stone]{2-5}
& $\ge 1000$
& 6944 (69.94)
& 5825 (75.50)
& 1119 (50.56) \\

\hline[stone]
\SetCell[c=2]{l}Sodium, mEq/L \\
\cline[stone]{2-5}
& $<$ 125
& 227 (2.29)
& 170 (2.20)
& 57 (2.58) \\
\cline[stone]{2-5}
& 125-144
& 8686 (87.49)
& 6890 (89.31)
& 1796 (81.16) \\
\cline[stone]{2-5}
& $\ge$ 145
& 1015 (10.22)
& 655 (8.49)
& 360 (16.27) \\

\hline[stone]
\SetCell[c=2]{l}Potassium \\
\cline[stone]{2-5}
& 3.0-4.9
& 7898 (79.55)
& 6350 (82.31)
& 1548 (69.95) \\
\cline[stone]{2-5}
& $<$ 3.0 or $\ge$ 5.0
& 2030 (20.45)
& 1365 (17.69)
& 665 (30.05) \\

\hline[stone]
\SetCell[c=2]{l}Bicarbonate, mEq/L \\
\cline[stone]{2-5}
& $<$ 15
& 729 (7.34)
& 364 (4.72)
& 365 (16.49) \\
\cline[stone]{2-5}
& 15-19
& 1977 (19.91)
& 1396 (18.09)
& 581 (26.25) \\
\cline[stone]{2-5}
& $\ge$ 20
& 7222 (72.74)
& 5955 (77.19)
& 1267 (57.25) \\

\hline[stone]
\SetCell[c=2]{l}Bilirubin, mg/dL \\
\cline[stone]{2-5}
& $<$ 4.0
& 9453 (95.22)
& 7456 (96.64)
& 1997 (90.24) \\
\cline[stone]{2-5}
& 4.0-5.9
& 157 (1.58)
& 96 (1.24)
& 61 (2.76) \\
\cline[stone]{2-5}
& $\ge$ 6.0
& 318 (3.20)
& 163 (2.11)
& 155 (7.00) \\

\hline[stone]
\SetCell[c=2]{l}White blood count, $\times 10^3$/mm$^3$ \\
\cline[stone]{2-5}
& $<$ 1.0
& 81 (0.82)
& 35 (0.45)
& 46 (2.08) \\
\cline[stone]{2-5}
& 1.0-19.9
& 8602 (86.64)
& 6896 (89.38)
& 1706 (77.09) \\
\cline[stone]{2-5}
& $\ge$ 20.0
& 1245 (12.54)
& 784 (10.16)
& 461 (20.83) \\

\hline[stone]
\SetCell[c=2]{l}Glasgow coma scale \\
\cline[stone]{2-5}
& 14-15
& 7590 (76.45)
& 6072 (78.70)
& 1518 (68.59) \\
\cline[stone]{2-5}
& 11-13
& 1142 (11.50)
& 900 (11.67)
& 242 (10.94) \\
\cline[stone]{2-5}
& 9-10
& 406 (4.09)
& 285 (3.69)
& 121 (5.47) \\
\cline[stone]{2-5}
& 6-8
& 455 (4.58)
& 297 (3.85)
& 158 (7.14) \\
\cline[stone]{2-5}
& $<$ 6
& 335 (3.37)
& 161 (2.09)
& 174 (7.86) \\

\hline[stone]
\SetCell[c=2]{l}Chronic disease \\
\cline[stone]{2-5}
& None
& 8682 (87.45)
& 6987 (90.56)
& 1695 (76.59) \\
\cline[stone]{2-5}
& Metastatic cancer
& 820 (8.26)
& 473 (6.13)
& 347 (15.68) \\
\cline[stone]{2-5}
& Hematologic malignancy
& 341 (3.43)
& 185 (2.40)
& 156 (7.05) \\
\cline[stone]{2-5}
& AIDS
& 85 (0.86)
& 70 (0.91)
& 15 (0.68) \\

\hline[stone]
\SetCell[c=2]{l}Type of admission \\
\cline[stone]{2-5}
& Scheduled surgical
& 60 (0.60)
& 58 (0.75)
& 2 (0.09) \\
\cline[stone]{2-5}
& Medical
& 8358 (84.19)
& 6442 (83.50)
& 1916 (86.58) \\
\cline[stone]{2-5}
& Unscheduled surgical
& 1510 (15.21)
& 1215 (15.75)
& 295 (13.33) \\
\end{longtblr}
}

\paragraph{NYP/WCM fracture cohort.}

We assembled a retrospective cohort from NewYork-Presbyterian/Weill Cornell Medicine electronic health records (Table \ref{tab:fracture}). Eligible patients were aged 50 years or older, underwent at least one DXA scan between January 1, 2012 and September 3, 2025, and had at least two years of follow-up. Demographics, prior fracture status, comorbidities, medication exposures, and DXA-derived T-scores were extracted from the electronic health record. T-scores above 1 were excluded as likely artifacts or documentation errors. FRAX score \cite{Kanis2008FRAX} was also extracted from the EHR.

For survival analysis, each patient’s earliest DXA scan was defined as baseline. The outcome was an incident fragility fracture occurring at least 15 days after baseline; fractures within 15 days were treated as prior events. Patients were followed from baseline until an incident fracture or the last documented clinical encounter. The final cohort comprised 11,522 patients, of whom 859 experienced a fracture during follow-up.

{
\definecolor{stone}{HTML}{BFBA9C}
\definecolor{grey}{HTML}{F6F3EB}
\footnotesize
\begin{longtblr}[
    caption = {Participant demographic and clinical characteristics on the NYP/WCM fracture cohort.},
    label = {tab:fracture},
]{
    colspec={lXlll},
    rowhead = 2,
    hline{1,3,Z} = {0.08em, black},
}
&&\SetCell[c=3]{l}{\textbf{Participants, No. (\%)}}\\
\cline[black]{3-5}
\SetCell[c=2]{l}\textbf{Characteristic} & & \textbf{All (N=11522)} & \textbf{Fracture (n=859)} & \textbf{No Fracture (n=10663)} \\
\SetCell[c=2]{l}Age, mean (SD), y && 68.40 (8.75) & 72.03 (8.88) & 68.11  (8.68) \\
\hline[stone]
\SetCell[c=2]{l}Body mass index, mean (SD), kg/m$^2$ && 25.37 (5.12) & 25.48 (5.34) & 25.37 (5.11) \\
\hline[stone]
\SetCell[c=2]{l}Sex\\
\cline[stone]{2-5}
& Female
  & 10527 (91.4) & 739 (86.0) & 9788 (91.8) \\
\cline[stone]{2-5}
& Male & 995 (8.6) & 120 (14.0) & 875 (8.2) \\
\hline[stone]
\SetCell[c=2]{l}Smoking status \\
\cline[stone]{2-5}
& Never smoker & 6777 (58.8) & 451 (52.5) & 6326 (59.3) \\
\cline[stone]{2-5}
& Former smoker & 2927 (25.4) & 240 (27.9) & 2687 (25.2) \\
\cline[stone]{2-5}
& Light smoker & 17 (0.1) & 0 (0.0) & 17 (0.2) \\
\cline[stone]{2-5}
& Moderate smoker & 570 (4.9) & 69 (8.0) & 501 (4.7) \\
\cline[stone]{2-5}
& Heavy smoker & 3 (0.0) & 0 (0.0) & 3 (0.0) \\
\cline[stone]{2-5}
& Current smoker & 27 (0.2) & 1 (0.1) & 26 (0.2) \\
\cline[stone]{2-5}
& Unknown & 1201 (10.4) & 98 (11.4) & 1103 (10.3) \\
\hline[stone]
\SetCell[c=2]{l}Alcohol use \\
\cline[stone]{2-5}
& Never drinker & 2808 (24.4) & 305 (35.5) & 2503 (23.5) \\
\cline[stone]{2-5}
& Former drinker & 13 (0.1) & 2 (0.2) & 11 (0.1) \\
\cline[stone]{2-5}
& Current drinker & 86 (0.7) & 9 (1.0) & 77 (0.7) \\
\cline[stone]{2-5}
& Unknown & 8615 (74.8) & 543 (63.2) & 8072 (75.7) \\
\hline[stone]
\SetCell[c=2]{l}Previous fracture && 961 (8.3) & 305 (35.5) & 656 (6.2) \\
\hline[stone]
\SetCell[c=2]{l}DXA Result\\
\cline[stone]{2-5}
& Lumbar spine (L1--L4) T-score, mean (SD)
  & -1.09 (1.02) & -1.23 (1.09) & -1.08 (1.01) \\
\cline[stone]{2-5}
& Femoral neck T-score, mean (SD)
  & -1.61 (0.62) & -1.80 (0.68) & -1.60 (0.62) \\
\cline[stone]{2-5}
& Total hip T-score, mean (SD)
  & -1.22 (0.76) & -1.51 (0.82) & -1.20 (0.76) \\
\hline[stone]
\SetCell[c=2]{l}Comorbidities \\
\cline[stone]{2-5}
& Alzheimer's disease & 76 (0.7) & 16 (1.9) & 60 (0.6) \\\cline[stone]{2-5}
& Ankylosing spondylitis & 18 (0.2) & 3 (0.3) & 15 (0.1) \\\cline[stone]{2-5}
& Asthma & 1391 (12.1) & 126 (14.7) & 1265 (11.9) \\\cline[stone]{2-5}
& Celiac disease & 131 (1.1) & 13 (1.5) & 118 (1.1) \\\cline[stone]{2-5}
& Cirrhosis & 244 (2.1) & 23 (2.7) & 221 (2.1) \\\cline[stone]{2-5}
& Chronic kidney disease & 654 (5.7) & 85 (9.9) & 569 (5.3) \\\cline[stone]{2-5}
& Chronic obstructive pulmonary disease & 442 (3.8) & 69 (8.0) & 373 (3.5) \\\cline[stone]{2-5}
& Crohn's disease & 153 (1.3) & 12 (1.4) & 141 (1.3) \\\cline[stone]{2-5}
& Cushing syndrome & 21 (0.2) & 4 (0.5) & 17 (0.2) \\\cline[stone]{2-5}
& Eating disorder & 54 (0.5) & 7 (0.8) & 47 (0.4) \\\cline[stone]{2-5}
& Hyperparathyroidism & 638 (5.5) & 35 (4.1) & 603 (5.7) \\\cline[stone]{2-5}
& Hypogonadism & 157 (1.4) & 16 (1.9) & 141 (1.3) \\\cline[stone]{2-5}
& Menopause & 772 (6.7) & 36 (4.2) & 736 (6.9) \\\cline[stone]{2-5}
& Multiple myeloma & 125 (1.1) & 18 (2.1) & 107 (1.0) \\\cline[stone]{2-5}
& Parkinson's disease & 85 (0.7) & 20 (2.3) & 65 (0.6) \\\cline[stone]{2-5}
& Rheumatoid arthritis & 401 (3.5) & 46 (5.4) & 355 (3.3) \\\cline[stone]{2-5}
& Systemic lupus erythematosus & 109 (0.9) & 8 (0.9) & 101 (0.9) \\\cline[stone]{2-5}
& Stroke & 463 (4.0) & 71 (8.3) & 392 (3.7) \\\cline[stone]{2-5}
& Type 1 diabetes & 122 (1.1) & 11 (1.3) & 111 (1.0) \\\cline[stone]{2-5}
& Type 2 diabetes & 1737 (15.1) & 172 (20.0) & 1565 (14.7) \\\cline[stone]{2-5}
& Ulcerative colitis & 184 (1.6) & 20 (2.3) & 164 (1.5) \\
\hline[stone]
\SetCell[c=2]{l}Medications \\
\cline[stone]{2-5}
& Abaloparatide & 139 (1.2) & 27 (3.1) & 112 (1.1) \\\cline[stone]{2-5}
& Antiandrogens & 34 (0.3) & 3 (0.3) & 31 (0.3) \\\cline[stone]{2-5}
& Antiretrovirals & 772 (6.7) & 61 (7.1) & 711 (6.7) \\\cline[stone]{2-5}
& Aromatase inhibitors & 1035 (9.0) & 55 (6.4) & 980 (9.2) \\\cline[stone]{2-5}
& Bisphosphonates & 1760 (15.3) & 179 (20.8) & 1581 (14.8) \\\cline[stone]{2-5}
& Calcineurin inhibitors & 1156 (10.0) & 104 (12.1) & 1052 (9.9) \\\cline[stone]{2-5}
& Denosumab & 506 (4.4) & 41 (4.8) & 465 (4.4) \\\cline[stone]{2-5}
& Enzyme-inducing antiepileptic drugs & 141 (1.2) & 19 (2.2) & 122 (1.1) \\\cline[stone]{2-5}
& Gonadotropin-releasing hormone agonist/antagonist & 281 (2.4) & 30 (3.5) & 251 (2.4) \\\cline[stone]{2-5}
& Romosozumab & 33 (0.3) & 3 (0.3) & 30 (0.3) \\\cline[stone]{2-5}
& Systemic steroids & 2650 (23.0) & 262 (30.5) & 2388 (22.4) \\\cline[stone]{2-5}
& Teriparatide & 139 (1.2) & 27 (3.1) & 112 (1.1) \\
\end{longtblr}
}

\subsection{Overview of \Ourmodel}

\begin{figure}[t]
    \centering
    \includegraphics[width=\linewidth]{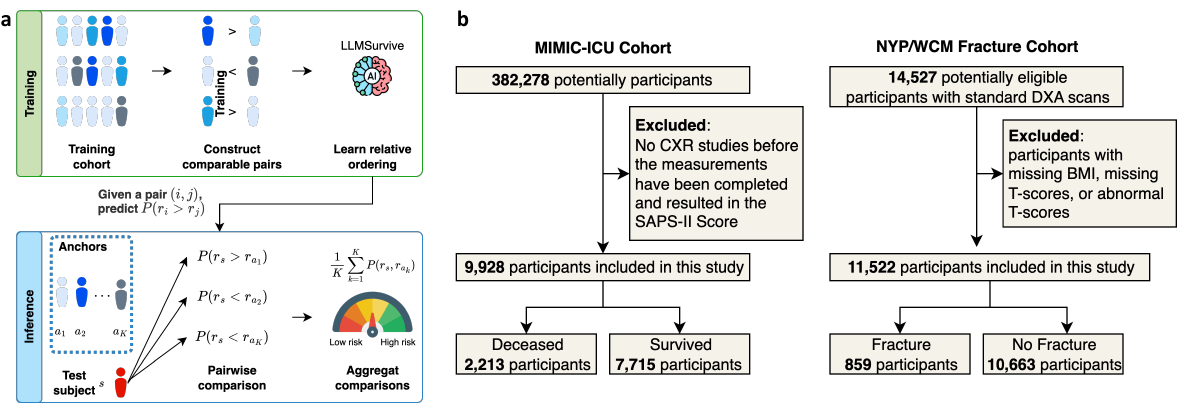}
    \caption{\textbf{\Ourmodel overview and study cohorts.}
\textbf{a}, Schematic of \Ourmodel. During training, the large language model is fine-tuned on comparable patient pairs to learn the relative ordering of survival times. During inference, each test patient is compared against a fixed set of anchor patients from the training cohort, and the aggregated outcomes define a final risk score.  
\textbf{b}, Cohort construction for the two clinical tasks: ICU mortality prediction in MIMIC-ICU and fracture risk assessment in the NYP/WCM fracture cohort.
}
\label{fig:overview}
\end{figure}

\Ourmodel comprises three stages (Figure~\ref{fig:overview}a). First, structured covariates are transformed into natural-language descriptions using a standardized template. Second, survival prediction is reformulated as a pairwise ranking task. The LLM is fine-tuned on comparable subject pairs to decide which subject is expected to experience the event earlier. 
Third, at inference, each test subject is compared with a set of training anchors, and the fraction of comparisons in which the test subject is ranked at higher risk is used as the final risk score (See Section \ref{sec:methods}: Methods).

This formulation offers two key advantages. First, it naturally incorporates all valid comparable pairs, including those involving censored observations, in a manner consistent with the rationale underlying the concordance index (C-index). Second, identifying which of two subjects experiences the event first is often more intuitive and better defined than directly estimating survival probabilities \cite{Qin2024-eo}. This pairwise perspective also aligns with the core assumption of the Cox proportional hazards model: that the relative risk, or hazard ratio, between subjects remains constant over time.

We used Llama 3.1-8B-Instruct as the primary backbone model and evaluated Qwen 2.5-7B-Instruct in parallel. We applied the same framework to two distinct clinical settings (Figure~\ref{fig:overview}b): ICU mortality prediction in MIMIC-IV \cite{johnson2020mimic-iv} and fragility fracture prediction in the NYP/WCM fracture cohort. These tasks span markedly different time horizons and disease dynamics, providing a stringent test of generalizability.


\subsection{Natural-language conversion of tabular variables}

To adapt structured tabular data for LLM input, each variable was converted into a natural-language statement using the template ``\texttt{\{column\} is \{value\}}'' \cite{hegselmann2023tabllm, fang2024large}. The covariates used for each cohort are listed in eTable \ref{covariate_icu} and eTable \ref{covariate_fracture} in Supplement \ref{sec:suppl}, respectively.

Statements were concatenated using semicolons. 
An example of the input is: ``Age is 79; Gender is FEMALE; T-score (NECK AVG) category is Osteopenia; Osteoporosis-related disease count is 0; Previous fracture is no''. Domain-specific terminology is explained at the beginning of the prompt to ensure interpretability. 

This approach preserves variable identity and value explicitly while presenting the input in a form directly consumable by the language model \cite{dinh2022lift}. 

\subsection{Pairwise ranking formulation for censored survival analysis}

Each subject was represented as a tuple \((x_i, e_i, t_i)\), where \(x_i\) denotes the natural-language input, \(e_i \in \{0,1\}\) indicates event occurrence and \(t_i\) denotes event or censoring time. Training pairs were constructed from comparable pairs \((i,j)\), defined as those in which subject \(i\) experienced the event (\(e_i = 1\)) and did so before subject \(j\)’s follow-up time:
\begin{equation}
\mathcal{P} = \{(i,j)\mid e_i = 1,\ t_i < t_j\}.
\end{equation}
The construction of pairs is covariate agnostic and depends only on the event or censoring time and event status label. For each pair, the model was trained to predict which subject would experience the event earlier. 
This corresponds to learning relative risk ranking rather than absolute survival time. The probability assigned to the model’s chosen next token served as the comparison score $r$, and supervised fine-tuning optimized a binary cross-entropy objective over valid comparable pairs.
\begin{equation}
\mathcal{L}_{\text{rank}} = - \sum_{(i,j) \in \mathcal{P}} \log P(r_i > r_j).
\end{equation}
Because the number of valid pairs can be large, we used a \textbf{nested case-control sampling strategy} \cite{Liddell1977-dm, Goldstein1992-gf}. For each event case, we randomly selected $N$ subjects with longer follow-up times. The optimal value of $N$ was determined by the model's performance on the development set.

Table \ref{tab:dataset_sampling} lists the training set size, the number of positive labels, the total number of possible pairs, the number of pairs remaining after nested case-control sampling, and the percentage of censored subjects in the created pairs. Within each pair, the first subject must have experienced an observed event, because a definitive event time is necessary to establish a valid comparison. The second subject may have either an observed event or a censoring time, provided that their time is after the event time of the first subject. The percentage of censored events in the final case-control pairs depends on the overall event rate in the cohort. For instance, this percentage is much higher in the NYP/WCM cohort since most of its samples are censored. Without case-control sampling, the total number of pairs would be computationally prohibitive for model training. After sampling, the dataset is reduced to a manageable size, while remaining an unbiased selection process and avoiding the pitfall of removing censored subjects.

{
\begin{table}[t]
\caption{Dataset characteristics and case-control sampling results.}
\label{tab:dataset_sampling}
\definecolor{stone}{HTML}{BFBA9C}
\definecolor{grey}{HTML}{F6F3EB}
\footnotesize
\begin{tblr}{
    colspec={lX[1]X[1]X[1.2]X[2.5]X[2.5]},
    rowhead = 1,
    hlines={stone},
    hline{1,2,Z} = {0.08em, black},
}
\textbf{Dataset} & \textbf{Training set size} & \textbf{Training set positive} & \textbf{Total possible pairs} & \textbf{Pairs after case-control sampling (N = 30)} & \textbf{Percentage of Censored Events in the Sampled Case-Control Pairs} \\
MIMIC-ICU & 6951 & 1542 & 6022023 & 46052 & 78.5\% \\
NYP/WCM Fracture & 8065 & 584 & 3023548 & 15720 & 95.7\% \\
\end{tblr}
\end{table}
}
\subsection{Prompt design}

For ICU mortality prediction, prompts described two patients and asked the model to decide which patient was expected to die first (Box~\ref{prompt:mimic}). For fracture prediction, prompts similarly described two patients and asked which was expected to sustain a fracture first, with brief explanations of domain-specific variables included at the beginning of the prompt (Box \ref{prompt:fracture}).
During training and testing, subject order within prompts was randomized to reduce positional bias.

\begin{prompt}[Prompt used for ICU mortality prediction on MIMIC-ICU]
\label{prompt:mimic}
You are a genius ICU specialist. You are given descriptions of two patients. Your task is to decide which of them is expected to decease first:

\begin{promptenum}
    \item {[}INSTANCE 1{]}
    \item {[}INSTANCE 2{]}
\end{promptenum}

Please provide your answer (\textbf{a} or \textbf{b}).

\end{prompt}

\begin{prompt}[Prompt used for fracture risk assessment on NYP/WCM Fracture]
\label{prompt:fracture}
You are a genius orthopedic specialist. Below are descriptions of the relevant terminologies:
\begin{promptitem}
    \item \textbf{Previous fracture}: Whether the patient has had a prior fracture.
    \item \textbf{Osteoporosis-related disease count}: The count of comorbidities that increase fracture risk (e.g., rheumatoid arthritis, COPD).
\end{promptitem}

You are given descriptions of two patients. Your task is to decide which of them is expected to sustain a fracture first:
\begin{promptenum}
\item {[}INSTANCE 1{]}
\item {[}INSTANCE 2{]}
\end{promptenum}

Please provide your answer (\textbf{a} or \textbf{b}).
\end{prompt}

\subsection{Inference via anchor comparison}

In inference, the risk for a test subject was estimated by comparison against a fixed set of anchor points. The anchors are randomly selected from the training set. For each cohort, they remain fixed once selected. Ablation studies regarding different anchor point selection methods and the number of anchors is discussed in Section \ref{anchor_ablation}. Let \(A = \{a_k\}_{k=1}^{K}\) denote the anchor set \cite{Liu2025-aa}. For each anchor \(a_k\), the model predicts whether the test subject $s$ would experience the event earlier than the anchor. 
This comparison then reduces to a binary decision using a threshold of 0.5:
\begin{equation}
\hat{y}(s,a_k) = 
\begin{cases}
1, & \text{if $P(r_s, r_{a_k}) > 0.5$}, \\
0, & \text{otherwise}.
\end{cases}
\end{equation}

The final risk score was then computed as the proportion of anchors for which the test subject was ranked as higher risk:
\begin{equation}
\hat{r}(s)=\frac{1}{K}\sum_{k=1}^{K}\hat{y}(s,a_k).
\end{equation}
In the primary experiments, we set \(K = 50\). A higher score indicates greater predicted risk.

\subsection{Model backbones and training}
\label{sec:implementation}

We evaluated two locally deployable instruction-tuned LLMs: Llama 3.1-8B-Instruct \cite{Grattafiori2024-rd} and Qwen 2.5-7B-Instruct  \cite{Qwen2024-mg}. Fine-tuning was performed on a single A100 GPU without parameter-efficient fine-tuning (PEFT). Based on development-set behavior, the ICU mortality model was fine-tuned for 1 epoch, and the fracture model for 2 epochs. During inference, the temperature was set to a near-zero value (i.e., 0.00001) to produce deterministic outputs. \Ourmodel’s anchor count and sampling ratio were selected by the validation ablation in Fig. \ref{fig:neg}. The held-out test set was used exclusively for final evaluation.

\subsection{Baseline methods}

We compared \Ourmodel against conventional and deep survival baselines, including the Cox proportional hazards model, Cox-nnet \cite{ching2018cox}, Deephit \cite{lee2018deephit}, and DeepSurv \cite{katzman2018deepsurv}. The Cox proportional hazards model is one of the most widely adopted statistical models for survival analysis. It models the hazard of an event as a log-linear function of covariates and uses partial likelihood to estimate model parameters while accounting for right-censored observations. DeepSurv and Cox-nnet extend the Cox model by replacing the linear predictor with a deep neural network to more flexibly capture nonlinear relationships between covariates and risk. In contrast, DeepHit is the only baseline that explicitly incorporates a ranking loss. It constructs comparable pairs of individuals in the same way as our work. For each pair, DeepHit compares their predicted cumulative incidence, defined as the predicted probability that the event has occurred by the earlier individual’s event time. It encourages the earlier individual to have a higher predicted cumulative incidence. Specifically, DeepHit applies an exponential penalty when this ordering is violated or insufficiently separated, with a larger penalty when the individual who experiences the event earlier is assigned a lower cumulative incidence. 

Hyperparameters for all methods were selected on the validation split. The details of the hyperparameter tuning for baseline models are listed in eTable \ref{tab:baseline_hyperparams}. All baselines and \Ourmodel were evaluated and tuned under the same train/validation/test splits, feature processing, and evaluation pipelines. We also compared against task-specific clinical scores: SAPS-II for ICU mortality \cite{leGall1993new} and FRAX for fracture risk \cite{Kanis2008FRAX, Kanis2009FRAXBone}.

\subsection{Evaluation metrics and statistical analysis}

Performance was assessed using the concordance index (C-index) and area under the receiver operating characteristic curve (AUC) at clinically relevant time horizons. The C-index was computed over all comparable pairs in the test set. AUC was calculated by thresholding predicted risk scores to classify event occurrence within each specified time horizon \cite{huang2005using}.

Confidence intervals were estimated using 1,000 bootstrap resamples at the patient level. Kaplan–Meier curves were generated on the test set after stratifying individuals into high- and low-risk groups using the median predicted risk from the training cohort.

\section{Results}

\subsection{\Ourmodel improves discrimination for ICU mortality prediction}

We first assess \Ourmodel on short-term ICU mortality prediction \cite{lipshutz2016predicting}. Prognostic scoring in critical care commonly relies on structured demographic, physiologic, and laboratory variables summarized by systems such as SAPS II \cite{leGall1993new}.

In MIMIC-ICU, \Ourmodel consistently outperformed conventional survival models (Figure~\ref{fig:main}a; eTable~\ref{tab_ICU_results} in Supplement \ref{sec:suppl}). Using Llama, \Ourmodel achieved an overall C-index of 0.776 (95\% confidence interval [CI], 0.752–0.800), compared with 0.753 (0.726–0.779) for the Cox proportional hazards model, 0.770 (0.744–0.796) for Cox-nnet, 0.730 (0.701–0.756) for DeepHit, and 0.766 (0.739–0.791) for DeepSurv. This corresponds to a 3.0 percentage-point improvement over the established clinical risk tool SAPS-II (0.746 [0.719–0.774]).
Using Qwen as the backbone yielded an overall C-index of 0.784 (0.761–0.807), indicating that the framework is robust across LLM families (eTable~\ref{tab_ICU_results} in Supplement \ref{sec:suppl}).
\begin{figure}[t]
    \centering
    \includegraphics[width=\linewidth]{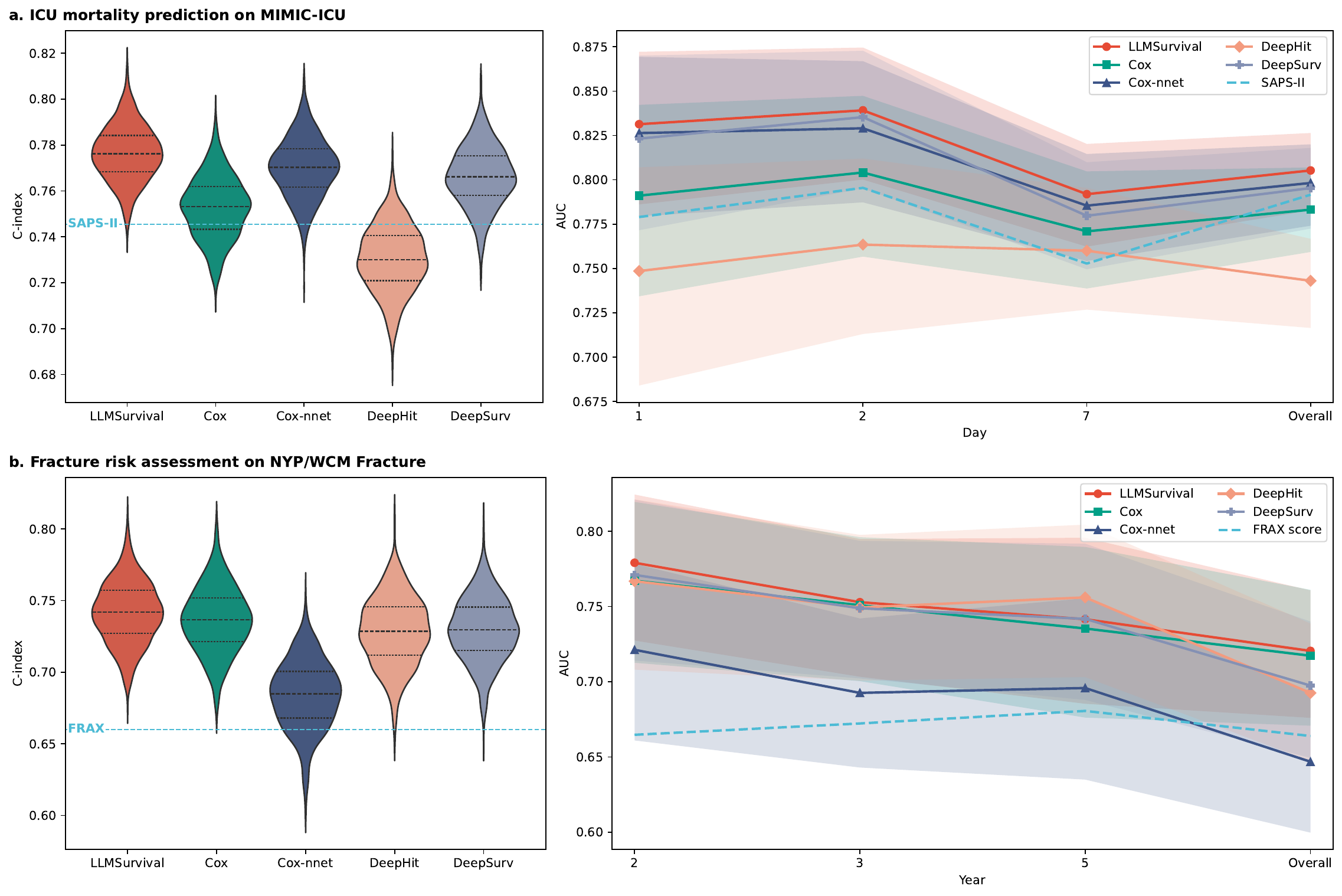}
    \caption{\textbf{Prognostic performance of \Ourmodel across acute and chronic clinical settings.}
    \textbf{a}, ICU mortality prediction on MIMIC-ICU. 
    \textbf{b}, Fracture risk assessment in the NYP/WCM fracture cohort.
    The left panels show distributions of overall C-index values across 1,000 patient-level bootstrap resamples for \Ourmodel, Cox, Cox-nnet, DeepHit, and DeepSurv, with SAPS-II or FRAX indicated by horizontal dashed lines. 
    The right panels show time-specific AUC values across clinically relevant follow-up horizons and overall evaluation. Shaded regions indicate 95\% confidence intervals.}
    \label{fig:main}
\end{figure}

Time-specific analyses showed similar trends. \Ourmodel achieved strong AUC values at 1 day, 2 days, 7 days, and overall, remaining above SAPS-II and competitive with or better than deep survival baselines across all horizons. 

These results suggest that the pairwise-ranking formulation captures clinically relevant short-term deterioration in critically ill patients more effectively than conventional survival modeling approaches trained on the same structured inputs.

\subsection{\Ourmodel shows robust overall performance for fracture risk prediction}

We next evaluated \Ourmodel for fragility fracture prediction using the NYP/WCM cohort (Figure~\ref{fig:main}b; eTable~\ref{tab_Fracture_results} in Supplement \ref{sec:suppl}). Fracture prediction is a longer-horizon task with lower event frequency and a distinct set of clinically relevant covariates, providing a complementary test of the framework \cite{GBDFracture2019, WHOFragility2024, USPSTF_OsteoporosisScreening_2025, NicholsonJAMA2025}. 

In this setting, \Ourmodel again achieved strong discrimination. Using Llama, \Ourmodel reached an overall C-index of 0.742 (0.699–0.783), compared with 0.737 (0.694–0.782) for Cox, 0.684 (0.632–0.728) for Cox-nnet, 0.729 (0.681–0.773) for DeepHit, and 0.730 (0.682–0.772) for DeepSurv. 
This corresponds to an 8.2 percentage-point improvement over the established clinical reference standard FRAX with a C-index of 0.660 (0.614–0.701) at year 10 \cite{Kanis2008FRAX, Kanis2009FRAXBone}. 

Time-specific AUC analysis showed that \Ourmodel consistently outperformed FRAX across 2-, 3-, and 5-year horizons and surpassed Cox and Cox-nnet at multiple time points. Performance was comparable to DeepSurv overall. Results obtained with Qwen were similar to those obtained with Llama, further supporting backbone robustness (eTable \ref{tab_Fracture_results} in Supplement \ref{sec:suppl}).

Together with the ICU results, these findings indicate that \Ourmodel generalizes across both acute and chronic clinical prediction tasks without task-specific architecture changes.

\subsection{\Ourmodel risk scores stratify patients into distinct survival groups}

To evaluate clinical usefulness beyond summary discrimination metrics, we examined Kaplan–Meier survival curves for groups defined by \Ourmodel-predicted risk. In each task, participants in the test set were split into high- and low-risk groups based on the median risk score derived from the training cohort.

\begin{figure}[t]
    \centering
    \includegraphics[width=0.7\linewidth]{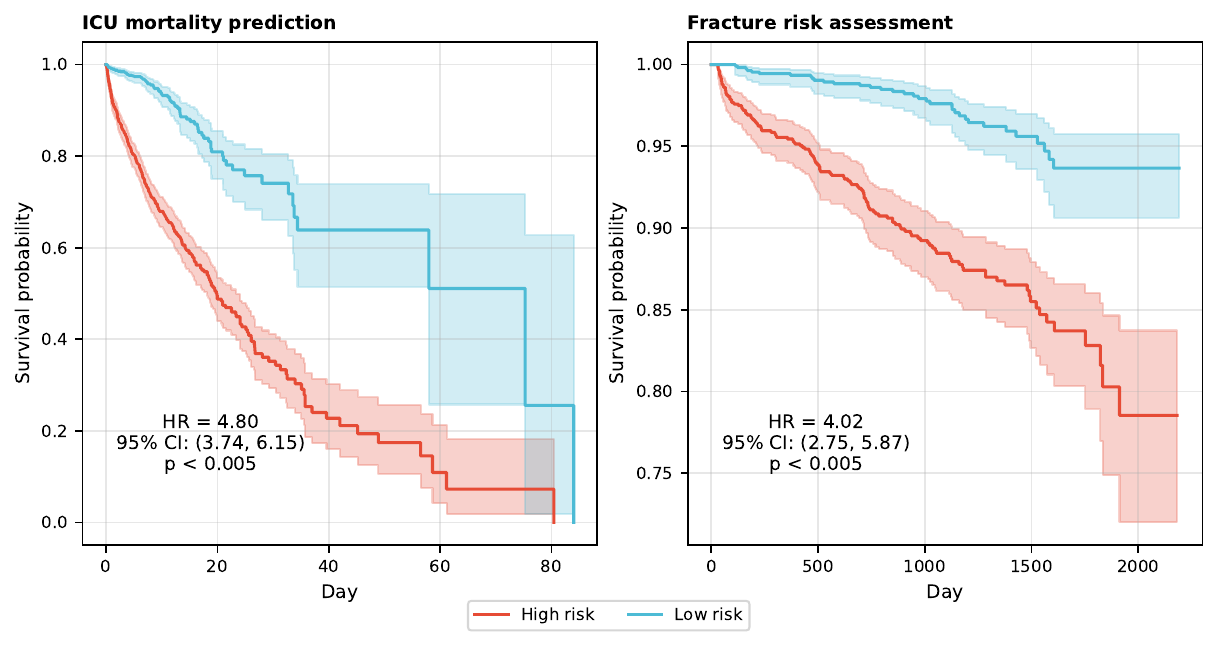}
    \caption{\textbf{Kaplan–Meier survival curves stratified by \Ourmodel risk score.} Participants were stratified into high- and low-risk groups using the median predicted risk score derived from the training cohort. Kaplan–Meier curves were computed on the held-out test set for ICU mortality prediction and fracture risk assessment. Hazard ratios, 95\% confidence intervals, and significance values are shown.
}
    \label{fig:3_4_KM}
\end{figure}

In both ICU mortality and fracture prediction, \Ourmodel produced clear separation between groups over time (Figure~\ref{fig:3_4_KM}). For ICU mortality, the hazard ratio comparing high-risk with low-risk groups was 4.80 (95\% CI, 3.74–6.15; $P < 0.005$). For fracture prediction, the hazard ratio was 4.02 (95\% CI, 2.75–5.87; $P < 0.005$). These results indicate that \Ourmodel risk scores define clinically meaningful strata with distinct event trajectories.

\subsection{Ablation studies}
\label{anchor_ablation}
\begin{figure}[t]
    \centering
    \begin{subfigure}[b]{0.49\textwidth}
    \includegraphics[width=\linewidth]{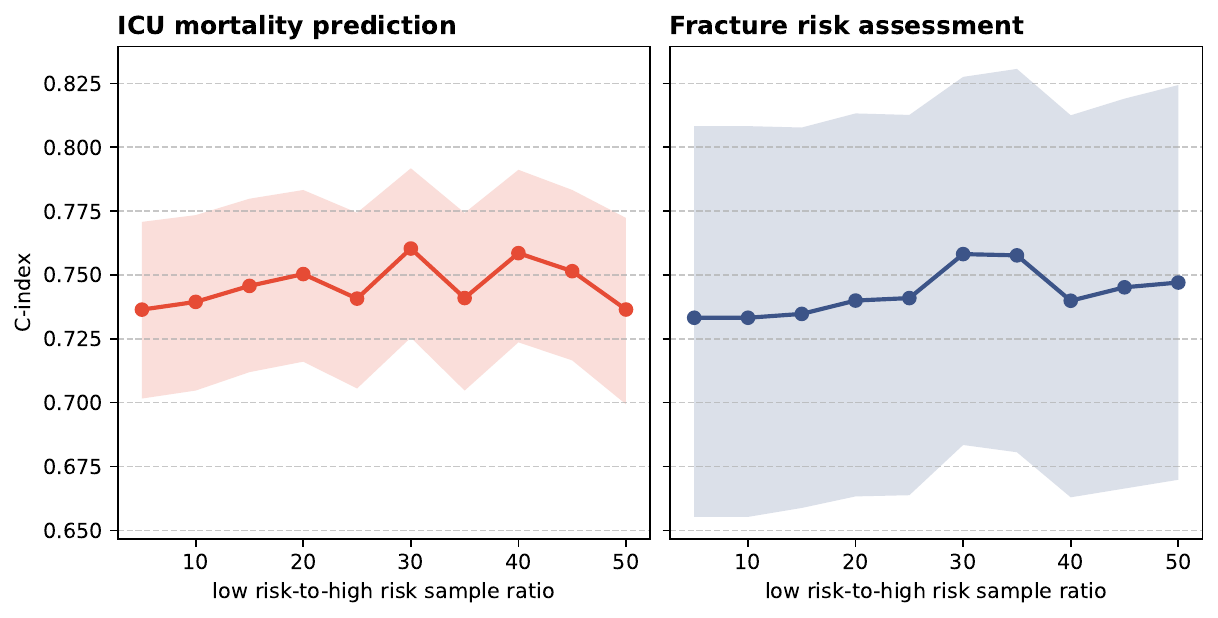}
    \caption{}
    \end{subfigure}
    \hfill
    \begin{subfigure}[b]{0.49\textwidth}
    \includegraphics[width=\linewidth]{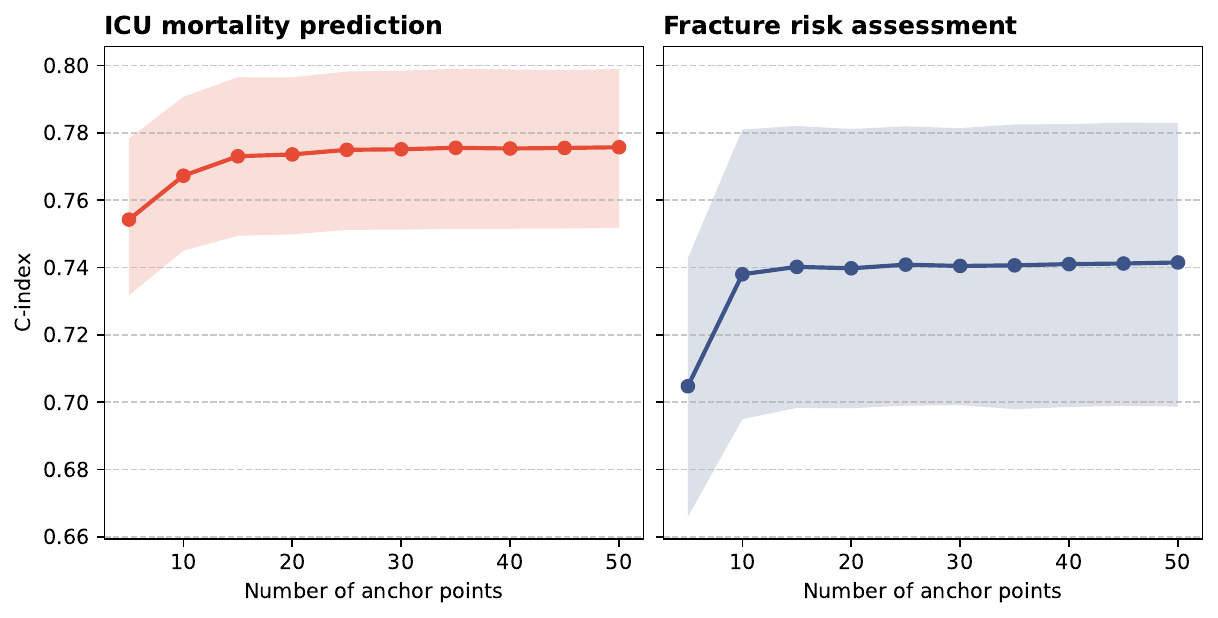}
    \caption{}
    \end{subfigure}
    \caption{\textbf{Effect of two implementation choices in \Ourmodel.} 
    \textbf{a}, Effect of the low-to-high risk ratio during model training. 
    \textbf{b}, Effect of anchor count during inference.
    Shaded bands indicate 95\% confidence intervals.}
    \label{fig:neg}
\end{figure}

We conducted several ablation studies to evaluate the impact of specific implementation details. First, we varied the ratio of low-risk to high-risk participants used during pairwise fine-tuning. Increasing the number of low-risk samples initially improved performance, indicating stronger ranking supervision, but gains plateaued and then declined at higher ratios (Figure~\ref{fig:neg}a). This pattern was observed across tasks, suggesting that moderate case-control sampling offers a favorable trade-off between signal and noise.

Second, we varied the number of anchor points used during inference. As expected, increasing anchor count improved the stability of risk estimation and increased C-index values initially, after which performance saturated (Figure~\ref{fig:neg}b). In the primary experiments, we used 50 anchors.

Thirdly, different anchor point selection strategies were evaluated. We explored two anchor selection strategies: the first, which is the current production setup, randomly samples anchors from all training instances. The second strategy randomly samples anchors only from instances with observed events. As shown in Supplementary  \ref{sec:anchor selection}, \Ourmodel is relatively robust to anchor point selection. 

Fourthly, we evaluated whether the risk score fluctuates drastically depending on the selected anchors. A hundred different anchor sets were simulated per cohort for each selection strategy, yielding 100 risk scores for each subject. The average coefficient of variation is reported in eTable \ref{tab:avg_cv}. Furthermore, we pooled the risk scores produced by both strategies and calculated the overall coefficient of variation. As shown in the table, the coefficient of variation remains consistently around 0.14 across both cohorts, individual selection methods, and the pooled results. This relatively low variance indicates that the risk score remains stable regardless of the specific anchors chosen.

The fifth ablation study was to evaluate whether the performance gains stem from the LLM's inherent capabilities or from the proposed pairwise reformulation and anchor aggregation. Specifically, we replaced the LLM with a logistic regression model to predict which subject in a given pair has a higher risk. The results (eTable \ref{tab_ICU_results} and eTable \ref{tab_Fracture_results}) show that the logistic regression baseline performs significantly worse than \Ourmodel on the MIMIC-ICU dataset, and slightly worse on the WCM/NYP fracture dataset. These findings demonstrate that the performance improvement is not attributable solely to the pairwise and anchor-based design, but also reflects the strong ability of LLMs to evaluate relative risk, echoing previous research showing that LLMs can be competitive with traditional logistic regression approaches \citet{hegselmann2023tabllm}.

Last but not least, subject ordering is another common concern in ranking-based methods. Specifically, whether the model yields a different prediction when evaluating if Subject A ranks higher than Subject B versus if Subject B ranks higher than Subject A. As shown in eSection \ref{sec:order}, this discrepancy is minimal in \Ourmodel.

%

\subsection{Illustrative examples support clinically coherent comparative reasoning}

\begin{figure}[t]
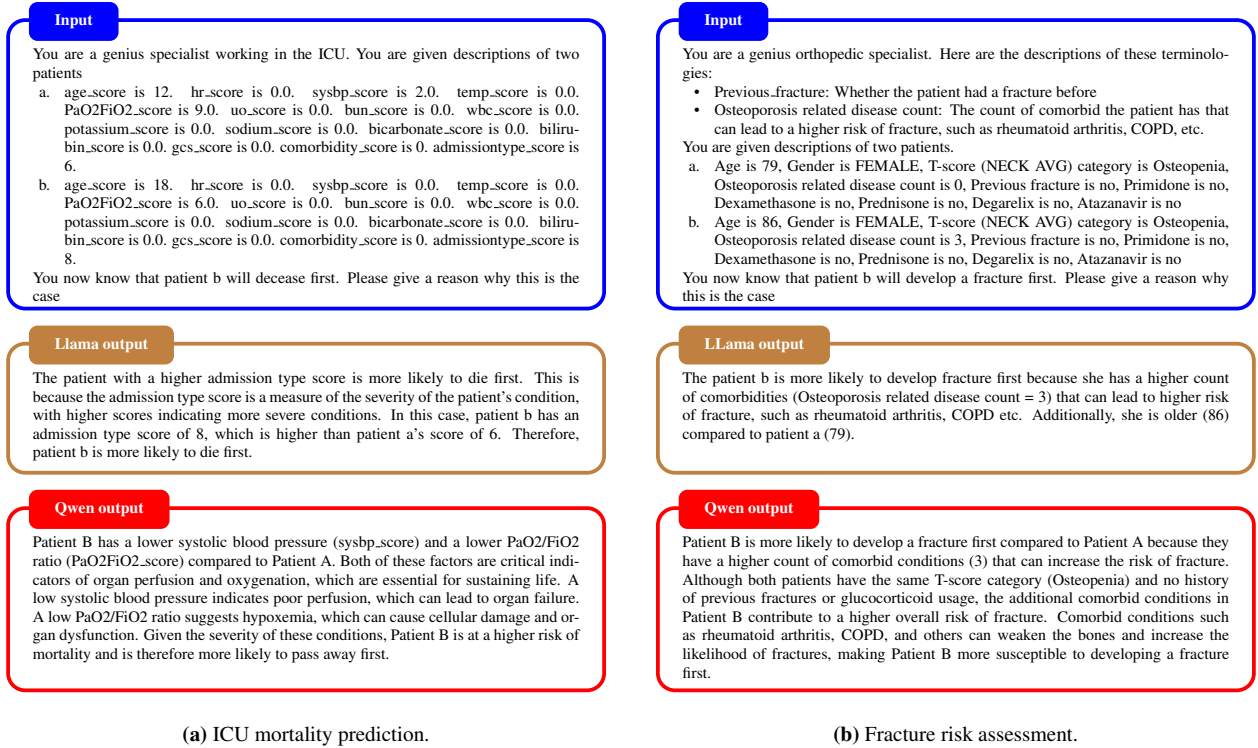

\tiny
\centering

\begin{minipage}[t]{0.48\textwidth}
     \begin{modelinput}[title=Input, height=110pt]
    You are a genius specialist working in the ICU. You are given descriptions of two patients
    \begin{enumerate}[label=\alph*.,nosep,leftmargin=2em]
    \item age\_score is 12. hr\_score is 0.0. sysbp\_score is 2.0. temp\_score is 0.0. PaO2FiO2\_score is 9.0. uo\_score is 0.0. bun\_score is 0.0. wbc\_score is 0.0. potassium\_score is 0.0. sodium\_score is 0.0. bicarbonate\_score is 0.0. bilirubin\_score is 0.0. gcs\_score is 0.0. comorbidity\_score is 0. admissiontype\_score is 6.
    \item age\_score is 18. hr\_score is 0.0. sysbp\_score is 0.0. temp\_score is 0.0. PaO2FiO2\_score is 6.0. uo\_score is 0.0. bun\_score is 0.0. wbc\_score is 0.0. potassium\_score is 0.0. sodium\_score is 0.0. bicarbonate\_score is 0.0. bilirubin\_score is 0.0. gcs\_score is 0.0. comorbidity\_score is 0. admissiontype\_score is 8.
    \end{enumerate}
    You now know that patient b will decease first. Please give a reason why this is the case 
    \end{modelinput}

    \begin{modeloutput}[title=Llama output,height=50pt]
    The patient with a higher admission type score is more likely to die first. This is because the admission type score is a measure of the severity of the patient's condition, with higher scores indicating more severe conditions. In this case, patient b has an admission type score of 8, which is higher than patient a's score of 6. Therefore, patient b is more likely to die first.
    \end{modeloutput}
    \begin{modeloutput2}[title=Qwen output,height=70pt]
    Patient B has a lower systolic blood pressure (sysbp\_score) and a lower PaO2/FiO2 ratio (PaO2FiO2\_score) compared to Patient A. Both of these factors are critical indicators of organ perfusion and oxygenation, which are essential for sustaining life. A low systolic blood pressure indicates poor perfusion, which can lead to organ failure. A low PaO2/FiO2 ratio suggests hypoxemia, which can cause cellular damage and organ dysfunction. Given the severity of these conditions, Patient B is at a higher risk of mortality and is therefore more likely to pass away first.
    \end{modeloutput2}

    \subcaption{ICU mortality prediction.}
\end{minipage}%
\hfill%
\begin{minipage}[t]{0.48\textwidth}
    \begin{modelinput}[title=Input, height=110pt]
    You are a genius orthopedic specialist. Here are the descriptions of these terminologies:
    \begin{itemize}[nosep,leftmargin=2em]
    \item Previous\_fracture: Whether the patient had a fracture before
    \item Osteoporosis related disease count: The count of comorbid the patient has that can lead to a higher risk of fracture, such as rheumatoid arthritis, COPD, etc.
    \end{itemize}
    
    You are given descriptions of two patients.
    \begin{enumerate}[label=\alph*.,nosep,leftmargin=2em]
    \item Age is 79, Gender is FEMALE, T-score (NECK AVG) category is Osteopenia, Osteoporosis related disease count is 0, Previous fracture is no, Primidone is no, Dexamethasone is no, Prednisone is no, Degarelix is no, Atazanavir is no
    \item Age is 86, Gender is FEMALE, T-score (NECK AVG) category is Osteopenia, Osteoporosis related disease count is 3, Previous fracture is no, Primidone is no, Dexamethasone is no, Prednisone is no, Degarelix is no, Atazanavir is no
    \end{enumerate}
    You now know that patient b will develop a fracture first. Please give a reason why this is the case
    \end{modelinput}
    \begin{modeloutput}[title=LLama output,height=50pt]
    The patient b is more likely to develop fracture first because she has a higher count of comorbidities (Osteoporosis related disease count = 3) that can lead to higher risk of fracture, such as rheumatoid arthritis, COPD etc. Additionally, she is older (86) compared to patient a (79).
    \end{modeloutput}
    \begin{modeloutput2}[title=Qwen output,height=70pt]
    Patient B is more likely to develop a fracture first compared to Patient A because they have a higher count of comorbid conditions (3) that can increase the risk of fracture. Although both patients have the same T-score category (Osteopenia) and no history of previous fractures or glucocorticoid usage, the additional comorbid conditions in Patient B contribute to a higher overall risk of fracture. Comorbid conditions such as rheumatoid arthritis, COPD, and others can weaken the bones and increase the likelihood of fractures, making Patient B more susceptible to developing a fracture first.
    \end{modeloutput2}

    \subcaption{Fracture risk assessment.}
\end{minipage}

\caption{\textbf{Illustrative examples of generated explanations for pairwise predictions.}
\textbf{a}, ICU mortality prediction. \textbf{b}, Fracture risk assessment. Outputs are shown for Llama- and Qwen-based models and highlight clinically meaningful differences between patient profiles.
}
\label{fig:reasoning}
\end{figure}

To examine whether pairwise predictions aligned with plausible clinical reasoning, we generated explanatory outputs for representative patient pairs using the base LLMs conditioned on the fine-tuned ranking outcomes (Figure \ref{fig:reasoning}). In ICU mortality prediction, the models highlighted clinically meaningful differences related to admission severity, oxygenation, and blood pressure. In fracture prediction, the models emphasized older age and a higher burden of osteoporosis-related comorbidity.

These examples do not establish causal interpretability, but they suggest that \Ourmodel's comparison-based outputs can be accompanied by coherent, human-readable rationales grounded in known clinical risk factors.

\newtcolorbox{modelinput}[1][]{%
    enhanced,
    colback=white, 
    colframe=blue, 
    colbacktitle=blue,
    boxsep=-1mm,
    attach boxed title to top left={xshift=0.3cm,yshift*=-3mm},
    rounded corners,arc=2mm,
    fonttitle=\bfseries,
    #1
}

\newtcolorbox{modeloutput}[1][]{%
    enhanced,
    before skip=2mm,after skip=2mm, 
    colback=white, 
    colframe=brown, 
    colbacktitle=brown,
    boxsep=-1mm,
    attach boxed title to top left={xshift=0.3cm,yshift*=-3mm},
    rounded corners,arc=2mm,
    fonttitle=\bfseries,
    #1
}

\newtcolorbox{modeloutput2}[1][]{%
    enhanced,
    before skip=2mm,after skip=2mm, 
    colback=white, 
    colframe=red, 
    colbacktitle=red,
    boxsep=-1mm,
    attach boxed title to top left={xshift=0.3cm,yshift*=-3mm},
    rounded corners,arc=2mm,
    fonttitle=\bfseries,
    #1
}

\section{Discussion}

We developed \Ourmodel, a framework that enables end-to-end survival analysis with LLMs while accounting for censoring. By converting time-to-event prediction into pairwise ranking over comparable subjects, \Ourmodel resolves a key barrier that has limited direct use of LLMs for survival modeling: the absence of explicit labels for censored observations. This formulation also aligns with a core strength of language models, namely comparative reasoning over textual descriptions.

The framework can be deployed locally because the base models used in our experiments (specifically 7B and 8B variants) are publicly available and relatively small. This architecture mitigates privacy risks by eliminating the need for external API calls.

Across two clinical tasks with markedly different time scales and event profiles, the framework achieved robust overall discrimination and exceeded established clinical scores. In ICU mortality prediction, it outperformed Cox proportional hazards modeling and SAPS-II. In fracture prediction, it exceeded Cox and FRAX and remained competitive with deep survival models. These results suggest that LLMs can function as the primary model for structured survival prediction rather than solely as feature encoders.

An important feature of the framework is its portability. The same general pipeline was applied to an acute critical care task and a chronic fracture risk task without architectural customization. Importantly, \Ourmodel achieves discrimination exceeding established clinical risk scores developed through decades of expert-driven design (SAPS-II for ICU mortality and FRAX for fracture risk). These reference scores are based on manually curated variables and fixed weighting schemes. In contrast, \Ourmodel learns meaningful risk stratification solely from data, as evidenced by clear separation of high- and low-risk groups in the Kaplan–Meier analyses. These results highlight the clinical relevance of LLM-based survival models as flexible alternatives to traditional scoring systems.

More broadly, this work illustrates how reformulating a statistically challenging problem into a task aligned with LLM capabilities can extend the applicability of language models in biomedical machine learning.

This study has several limitations. First, training exhibited some variability across sampling configurations, indicating room for improved optimization stability. Second, pair selection was based on random sampling of comparable controls, and more structured sampling strategies may improve efficiency or performance. Third, inference requires multiple pairwise comparisons against anchors, which introduces additional computational cost. Finally, our experiments used structured tabular inputs only; future work should evaluate the integration of multimodal signals such as clinical notes and medical imaging.

\section{Conclusion}
In summary, \Ourmodel establishes a proof of concept for end-to-end censoring-aware survival analysis with LLMs. The findings support comparison-based reformulation as a practical route for adapting LLMs to censored clinical prediction tasks.

\section*{Reproducibility}

Both Llama- and Qwen-based versions of \Ourmodel showed similar trends across tasks. Additional analyses indicated that performance was relatively insensitive to anchor-selection strategy and robust to pair ordering.

\section*{Acknowledgements}

This research was supported by the National Library of Medicine under the grant numbers R01LM014306, R01LM014344, R01LM014573, R21EY035296 (Y.P.).

\section*{Ethics approval}
This study was approved by the Institutional Review Board of Weill Cornell Medicine (IRB Approval No. 2402027008)

\section*{Author contributions}
Study concepts/study design, Y.W., H.D., Y.P.; manuscript drafting or manuscript revision for important intellectual content, Y.W., H.D., Y.P.; agrees to ensure any questions related to the work are appropriately resolved, Y.W., H.D., Y.Lin, J.Q., Y.Liu, Y.P.; literature research, Y.W., H.D.; experimental studies, Y.W., H.D., Y.Lin; data interpretation and statistical analysis, Y.W., H.D., J.Q., Y.Liu; and manuscript editing, Y.W., H.D., Y.P.; read and approval of final version of the submitted manuscript, all authors.

\section*{Competing interests}\label{competing_interests}
None declared.

\section*{Data availability}
The MIMIC-ICU dataset used in this study is available from PhysioNet at \url{https://physionet.org/content/mimiciv}. 
Access to the NYP/WCM fracture cohort is subject to institutional review and data-sharing restrictions to protect patient privacy and comply with intellectual property obligations. Requests for academic use will be evaluated on a case-by-case basis and may require a material transfer agreement.

\section*{Code availablity}

Code supporting this study is available for academic research use at \url{https://github.com/bionlplab/llmsurvival}.

\newpage

\setlength{\bibsep}{3pt plus 0.3ex}
\bibliographystyle{unsrtnat}
\bibliography{ref,peng}

@MISC{johnson2020mimic-iv,
  title     = "{MIMIC}-{IV}",
  author    = "Johnson, Alistair and Bulgarelli, Lucas and Pollard, Tom and
               Horng, Steven and Celi, Leo Anthony and Mark, Roger",
  publisher = "PhysioNet",
  year      =  2020,
  doi       = "10.13026/A3WN-HQ05"
}

@ARTICLE{fang2024large,
  title   = "Large Language Models ({LLMs}) on Tabular Data: Prediction,
             Generation, and Understanding - A Survey",
  author  = "Fang, Xi and Xu, Weijie and Tan, Fiona Anting and Hu, Ziqing and
             Zhang, Jiani and Qi, Yanjun and Sengamedu, Srinivasan H and
             Faloutsos, Christos",
  journal = "Transactions on Machine Learning Research",
  year    =  2024,
  issn    = "2835-8856"
}

@INPROCEEDINGS{chao2024make,
  title     = "Make large language model a better ranker",
  author    = "Chao, Wen-Shuo and Zheng, Zhi and Zhu, Hengshu and Liu, Hao",
  booktitle = "Findings of the Association for Computational Linguistics: EMNLP
               2024",
  publisher = "Association for Computational Linguistics",
  address   = "Stroudsburg, PA, USA",
  pages     = "918--929",
  month     =  nov,
  year      =  2024,
  doi       = "10.18653/v1/2024.findings-emnlp.51"
}

@ARTICLE{chung1991survival,
  title     = "Survival analysis: A survey",
  author    = "Chung, Ching-Fan and Schmidt, Peter and Witte, Ana D",
  journal   = "J. Quant. Criminol.",
  publisher = "Springer Science and Business Media LLC",
  volume    =  7,
  number    =  1,
  pages     = "59--98",
  month     =  mar,
  year      =  1991,
  doi       = "10.1007/bf01083132",
  issn      = "0748-4518,1573-7799"
}

@ARTICLE{huang2005using,
  title     = "Using {AUC} and accuracy in evaluating learning algorithms",
  author    = "Huang, Jin and Ling, C X",
  journal   = "IEEE Trans. Knowl. Data Eng.",
  publisher = "Institute of Electrical and Electronics Engineers (IEEE)",
  volume    =  17,
  number    =  3,
  pages     = "299--310",
  month     =  mar,
  year      =  2005,
  doi       = "10.1109/tkde.2005.50",
  issn      = "1041-4347,1558-2191"
}

@INPROCEEDINGS{hegselmann2023tabllm,
  title     = "{TabLLM}: Few-shot Classification of Tabular Data with Large
               Language Models",
  author    = "Hegselmann, Stefan and Buendia, Alejandro and Lang, Hunter and
               Agrawal, Monica and Jiang, Xiaoyi and Sontag, David",
  booktitle = "International Conference on Artificial Intelligence and
               Statistics",
  publisher = "PMLR",
  pages     = "5549--5581",
  month     =  "11~" # apr,
  year      =  2023,
  issn      = "2640-3498"
}

@INPROCEEDINGS{steinberg2023motor,
  title     = "{MOTOR}: A Time-to-Event Foundation Model For Structured Medical
               Records",
  author    = "Steinberg, Ethan and Fries, Jason Alan and Xu, Yizhe and Shah,
               Nigam",
  booktitle = "The Twelfth International Conference on Learning Representations",
  month     =  "13~" # oct,
  year      =  2023
}

@INPROCEEDINGS{dinh2022lift,
  title     = "{LIFT}: language-interfaced fine-tuning for non-language machine
               learning tasks",
  author    = "Dinh, Tuan and Zeng, Yuchen and Zhang, Ruisu and Lin, Ziqian and
               Gira, Michael and Rajput, Shashank and Sohn, Jy-Yong and
               Papailiopoulos, Dimitris and Lee, Kangwook",
  booktitle = "Proceedings of the 36th International Conference on Neural
               Information Processing Systems",
  publisher = "Curran Associates Inc.",
  address   = "Red Hook, NY, USA",
  number    = "Article 855",
  pages     = "11763--11784",
  series    = "NIPS '22",
  month     =  "28~" # nov,
  year      =  2022,
  doi       = "10.5555/3600270.3601125",
  isbn      =  9781713871088
}

@ARTICLE{wang2019machine,
  title     = "Machine learning for survival analysis: A survey",
  author    = "Wang, Ping and Li, Yan and Reddy, Chandan K",
  journal   = "ACM Comput. Surv.",
  publisher = "Association for Computing Machinery (ACM)",
  volume    =  51,
  number    =  6,
  pages     = "1--36",
  month     =  "30~" # nov,
  year      =  2019,
  doi       = "10.1145/3214306",
  issn      = "0360-0300,1557-7341"
}

@ARTICLE{Johnson2023-sr,
  title     = "{MIMIC}-{IV}, a freely accessible electronic health record
               dataset",
  author    = "Johnson, Alistair E W and Bulgarelli, Lucas and Shen, Lu and
               Gayles, Alvin and Shammout, Ayad and Horng, Steven and Pollard,
               Tom J and Hao, Sicheng and Moody, Benjamin and Gow, Brian and
               Lehman, Li-Wei H and Celi, Leo A and Mark, Roger G",
  journal   = "Sci. Data",
  publisher = "Springer Science and Business Media LLC",
  volume    =  10,
  number    =  1,
  pages     =  1,
  month     =  "3~" # jan,
  year      =  2023,
  doi       = "10.1038/s41597-022-01899-x",
  pmc       = "PMC9810617",
  pmid      =  36596836,
  issn      = "2052-4463"
}

@ARTICLE{Zeng2024-ye,
  title         = "{LLM}-{RankFusion}: Mitigating intrinsic inconsistency in
                   {LLM}-based ranking",
  author        = "Zeng, Yifan and Tendolkar, Ojas and Baartmans, Raymond and
                   Wu, Qingyun and Chen, Lizhong and Wang, Huazheng",
  journal       = "arXiv [cs.IR]",
  month         =  "31~" # may,
  year          =  2024,
  archivePrefix = "arXiv",
  primaryClass  = "cs.IR",
  eprint        = "2406.00231"
}

@ARTICLE{Jing2019-nj,
  title     = "A deep survival analysis method based on ranking",
  author    = "Jing, Bingzhong and Zhang, Tao and Wang, Zixian and Jin, Ying and
               Liu, Kuiyuan and Qiu, Wenze and Ke, Liangru and Sun, Ying and He,
               Caisheng and Hou, Dan and Tang, Linquan and Lv, Xing and Li,
               Chaofeng",
  journal   = "Artif. Intell. Med.",
  publisher = "Elsevier BV",
  volume    =  98,
  pages     = "1--9",
  month     =  jul,
  year      =  2019,
  doi       = "10.1016/j.artmed.2019.06.001",
  pmid      =  31521247,
  issn      = "0933-3657,1873-2860"
}

@ARTICLE{Qwen2024-mg,
  title         = "{Qwen2}.5 Technical Report",
  author        = "{Qwen} and Yang, An and Yang, Baosong and Zhang, Beichen and
                   Hui, Binyuan and Zheng, Bo and Yu, Bowen and Li, Chengyuan
                   and Liu, Dayiheng and Huang, Fei and Wei, Haoran and Lin,
                   Huan and Yang, Jian and Tu, Jianhong and Zhang, Jianwei and
                   Yang, Jianxin and Yang, Jiaxi and Zhou, Jingren and Lin,
                   Junyang and Dang, Kai and Lu, Keming and Bao, Keqin and Yang,
                   Kexin and Yu, Le and Li, Mei and Xue, Mingfeng and Zhang, Pei
                   and Zhu, Qin and Men, Rui and Lin, Runji and Li, Tianhao and
                   Tang, Tianyi and Xia, Tingyu and Ren, Xingzhang and Ren,
                   Xuancheng and Fan, Yang and Su, Yang and Zhang, Yichang and
                   Wan, Yu and Liu, Yuqiong and Cui, Zeyu and Zhang, Zhenru and
                   Qiu, Zihan",
  journal       = "arXiv [cs.CL]",
  month         =  "19~" # dec,
  year          =  2024,
  archivePrefix = "arXiv",
  primaryClass  = "cs.CL",
  eprint        = "2412.15115"
}

@INPROCEEDINGS{Qin2024-eo,
  title     = "Large language models are effective text rankers with pairwise
               ranking prompting",
  author    = "Qin, Zhen and Jagerman, Rolf and Hui, Kai and Zhuang, Honglei and
               Wu, Junru and Yan, Le and Shen, Jiaming and Liu, Tianqi and Liu,
               Jialu and Metzler, Donald and Wang, Xuanhui and Bendersky,
               Michael",
  booktitle = "Findings of the Association for Computational Linguistics: NAACL
               2024",
  publisher = "Association for Computational Linguistics",
  address   = "Stroudsburg, PA, USA",
  year      =  2024,
  doi       = "10.18653/v1/2024.findings-naacl.97"
}

@INPROCEEDINGS{esteban2024predictive,
  title     = "Predictive maintenance with large language models and
               transformer-based survival analysis",
  author    = "Esteban, Aurora and Cobilean, Victor and Nawaratne, Rashmika",
  booktitle = "IECON 2024 - 50th Annual Conference of the IEEE Industrial
               Electronics Society",
  publisher = "IEEE",
  pages     = "1--6",
  month     =  "3~" # nov,
  year      =  2024,
  doi       = "10.1109/iecon55916.2024.10905382",
  isbn      = "9781665464543,9781665464550"
}

@ARTICLE{Liddell1977-dm,
  title     = "Methods of cohort analysis: Appraisal by application to asbestos
               mining",
  author    = "Liddell, F D K and McDonald, J C and Thomas, D C and Cunliffe,
               Stella V",
  journal   = "J. R. Stat. Soc. Ser. A",
  publisher = "JSTOR",
  volume    =  140,
  number    =  4,
  pages     =  469,
  year      =  1977,
  doi       = "10.2307/2345280",
  issn      = "0035-9238,2397-2327"
}

@ARTICLE{Lin2021-ye,
  title   = "An empirical study of using radiology reports and images to improve
             {ICU}-mortality prediction",
  author  = "Lin, Mingquan and Wang, Song and Ding, Ying and Zhao, Lihui and
             Wang, Fei and Peng, Yifan",
  journal = "IEEE Int. Conf. Healthc. Inform.",
  volume  =  2021,
  pages   = "497--498",
  month   =  aug,
  year    =  2021,
  doi     = "10.1109/ichi52183.2021.00088",
  pmc     = "PMC9076267",
  pmid    =  35531070
}

@ARTICLE{Goldstein1992-gf,
  title     = "Asymptotic theory for nested case-control sampling in the cox
               regression model",
  author    = "Goldstein, Larry and Langholz, Bryan",
  journal   = "Ann. Stat.",
  publisher = "Institute of Mathematical Statistics",
  volume    =  20,
  number    =  4,
  pages     = "1903--1928",
  month     =  "1~" # dec,
  year      =  1992,
  doi       = "10.1214/aos/1176348895",
  issn      = "0090-5364,2168-8966"
}

@INPROCEEDINGS{Liu2025-aa,
  title     = "{SurvUnc}: A meta-model based uncertainty quantification
               framework for survival analysis",
  author    = "Liu, Yu and Tao, Weiyao and Xia, Tong and Knight, Simon and Zhu,
               Tingting",
  booktitle = "Proceedings of the 31st ACM SIGKDD Conference on Knowledge
               Discovery and Data Mining V.2",
  publisher = "ACM",
  address   = "New York, NY, USA",
  pages     = "1903--1914",
  month     =  "3~" # aug,
  year      =  2025,
  doi       = "10.1145/3711896.3737140"
}

@ARTICLE{Van-Belle2011-ab,
  title     = "Support vector methods for survival analysis: a comparison
               between ranking and regression approaches",
  author    = "Van Belle, Vanya and Pelckmans, Kristiaan and Van Huffel, Sabine
               and Suykens, Johan A K",
  journal   = "Artif. Intell. Med.",
  publisher = "Elsevier BV",
  volume    =  53,
  number    =  2,
  pages     = "107--118",
  month     =  oct,
  year      =  2011,
  doi       = "10.1016/j.artmed.2011.06.006",
  pmid      =  21821401,
  issn      = "0933-3657,1873-2860"
}

@ARTICLE{Grattafiori2024-rd,
  title         = "The Llama 3 herd of models",
  author        = "Grattafiori, Aaron and Dubey, Abhimanyu and Jauhri, Abhinav
                   and Pandey, Abhinav and Kadian, Abhishek and Al-Dahle, Ahmad
                   and Letman, Aiesha and Mathur, Akhil and Schelten, Alan and
                   Vaughan, Alex and Yang, Amy and Fan, Angela and Goyal,
                   Anirudh and Hartshorn, Anthony and Yang, Aobo and Mitra,
                   Archi and Sravankumar, Archie and Korenev, Artem and
                   Hinsvark, Arthur and Rao, Arun and Zhang, Aston and
                   Rodriguez, Aurelien and Gregerson, Austen and Spataru, Ava
                   and Roziere, Baptiste and Biron, Bethany and Tang, Binh and
                   Chern, Bobbie and Caucheteux, Charlotte and Nayak, Chaya and
                   Bi, Chloe and Marra, Chris and McConnell, Chris and Keller,
                   Christian and Touret, Christophe and Wu, Chunyang and Wong,
                   Corinne and Ferrer, Cristian Canton and Nikolaidis, Cyrus and
                   Allonsius, Damien and Song, Daniel and Pintz, Danielle and
                   Livshits, Danny and Wyatt, Danny and Esiobu, David and
                   Choudhary, Dhruv and Mahajan, Dhruv and Garcia-Olano, Diego
                   and Perino, Diego and Hupkes, Dieuwke and Lakomkin, Egor and
                   AlBadawy, Ehab and Lobanova, Elina and Dinan, Emily and
                   Smith, Eric Michael and Radenovic, Filip and Guzm\'{a}n,
                   Francisco and Zhang, Frank and Synnaeve, Gabriel and Lee,
                   Gabrielle and Anderson, Georgia Lewis and Thattai, Govind and
                   Nail, Graeme and Mialon, Gregoire and Pang, Guan and
                   Cucurell, Guillem and Nguyen, Hailey and Korevaar, Hannah and
                   Xu, Hu and Touvron, Hugo and Zarov, Iliyan and Ibarra, Imanol
                   Arrieta and Kloumann, Isabel and Misra, Ishan and Evtimov,
                   Ivan and Zhang, Jack and Copet, Jade and Lee, Jaewon and
                   Geffert, Jan and Vranes, Jana and Park, Jason and Mahadeokar,
                   Jay and Shah, Jeet and van der Linde, Jelmer and Billock,
                   Jennifer and Hong, Jenny and Lee, Jenya and Fu, Jeremy and
                   Chi, Jianfeng and Huang, Jianyu and Liu, Jiawen and Wang, Jie
                   and Yu, Jiecao and Bitton, Joanna and Spisak, Joe and Park,
                   Jongsoo and Rocca, Joseph and Johnstun, Joshua and Saxe,
                   Joshua and Jia, Junteng and Alwala, Kalyan Vasuden and
                   Prasad, Karthik and Upasani, Kartikeya and Plawiak, Kate and
                   Li, Ke and Heafield, Kenneth and Stone, Kevin and El-Arini,
                   Khalid and Iyer, Krithika and Malik, Kshitiz and Chiu,
                   Kuenley and Bhalla, Kunal and Lakhotia, Kushal and
                   Rantala-Yeary, Lauren and van der Maaten, Laurens and Chen,
                   Lawrence and Tan, Liang and Jenkins, Liz and Martin, Louis
                   and Madaan, Lovish and Malo, Lubo and Blecher, Lukas and
                   Landzaat, Lukas and de Oliveira, Luke and Muzzi, Madeline and
                   Pasupuleti, Mahesh and Singh, Mannat and Paluri, Manohar and
                   Kardas, Marcin and Tsimpoukelli, Maria and Oldham, Mathew and
                   Rita, Mathieu and Pavlova, Maya and Kambadur, Melanie and
                   Lewis, Mike and Si, Min and Singh, Mitesh Kumar and Hassan,
                   Mona and Goyal, Naman and Torabi, Narjes and Bashlykov,
                   Nikolay and Bogoychev, Nikolay and Chatterji, Niladri and
                   Zhang, Ning and Duchenne, Olivier and \c{C}elebi, Onur and
                   Alrassy, Patrick and Zhang, Pengchuan and Li, Pengwei and
                   Vasic, Petar and Weng, Peter and Bhargava, Prajjwal and
                   Dubal, Pratik and Krishnan, Praveen and Koura, Punit Singh
                   and Xu, Puxin and He, Qing and Dong, Qingxiao and Srinivasan,
                   Ragavan and Ganapathy, Raj and Calderer, Ramon and Cabral,
                   Ricardo Silveira and Stojnic, Robert and Raileanu, Roberta
                   and Maheswari, Rohan and Girdhar, Rohit and Patel, Rohit and
                   Sauvestre, Romain and Polidoro, Ronnie and Sumbaly, Roshan
                   and Taylor, Ross and Silva, Ruan and Hou, Rui and Wang, Rui
                   and Hosseini, Saghar and Chennabasappa, Sahana and Singh,
                   Sanjay and Bell, Sean and Kim, Seohyun Sonia and Edunov,
                   Sergey and Nie, Shaoliang and Narang, Sharan and Raparthy,
                   Sharath and Shen, Sheng and Wan, Shengye and Bhosale, Shruti
                   and Zhang, Shun and Vandenhende, Simon and Batra, Soumya and
                   Whitman, Spencer and Sootla, Sten and Collot, Stephane and
                   Gururangan, Suchin and Borodinsky, Sydney and Herman, Tamar
                   and Fowler, Tara and Sheasha, Tarek and Georgiou, Thomas and
                   Scialom, Thomas and Speckbacher, Tobias and Mihaylov, Todor
                   and Xiao, Tong and Karn, Ujjwal and Goswami, Vedanuj and
                   Gupta, Vibhor and Ramanathan, Vignesh and Kerkez, Viktor and
                   Gonguet, Vincent and Do, Virginie and Vogeti, Vish and
                   Albiero, V\'{\i}tor and Petrovic, Vladan and Chu, Weiwei and
                   Xiong, Wenhan and Fu, Wenyin and Meers, Whitney and Martinet,
                   Xavier and Wang, Xiaodong and Wang, Xiaofang and Tan,
                   Xiaoqing Ellen and Xia, Xide and Xie, Xinfeng and Jia, Xuchao
                   and Wang, Xuewei and Goldschlag, Yaelle and Gaur, Yashesh and
                   Babaei, Yasmine and Wen, Yi and Song, Yiwen and Zhang, Yuchen
                   and Li, Yue and Mao, Yuning and Coudert, Zacharie Delpierre
                   and Yan, Zheng and Chen, Zhengxing and Papakipos, Zoe and
                   Singh, Aaditya and Srivastava, Aayushi and Jain, Abha and
                   Kelsey, Adam and Shajnfeld, Adam and Gangidi, Adithya and
                   Victoria, Adolfo and Goldstand, Ahuva and Menon, Ajay and
                   Sharma, Ajay and Boesenberg, Alex and Baevski, Alexei and
                   Feinstein, Allie and Kallet, Amanda and Sangani, Amit and
                   Teo, Amos and Yunus, Anam and Lupu, Andrei and Alvarado,
                   Andres and Caples, Andrew and Gu, Andrew and Ho, Andrew and
                   Poulton, Andrew and Ryan, Andrew and Ramchandani, Ankit and
                   Dong, Annie and Franco, Annie and Goyal, Anuj and Saraf,
                   Aparajita and Chowdhury, Arkabandhu and Gabriel, Ashley and
                   Bharambe, Ashwin and Eisenman, Assaf and Yazdan, Azadeh and
                   James, Beau and Maurer, Ben and Leonhardi, Benjamin and
                   Huang, Bernie and Loyd, Beth and De Paola, Beto and
                   Paranjape, Bhargavi and Liu, Bing and Wu, Bo and Ni, Boyu and
                   Hancock, Braden and Wasti, Bram and Spence, Brandon and
                   Stojkovic, Brani and Gamido, Brian and Montalvo, Britt and
                   Parker, Carl and Burton, Carly and Mejia, Catalina and Liu,
                   Ce and Wang, Changhan and Kim, Changkyu and Zhou, Chao and
                   Hu, Chester and Chu, Ching-Hsiang and Cai, Chris and Tindal,
                   Chris and Feichtenhofer, Christoph and Gao, Cynthia and
                   Civin, Damon and Beaty, Dana and Kreymer, Daniel and Li,
                   Daniel and Adkins, David and Xu, David and Testuggine, Davide
                   and David, Delia and Parikh, Devi and Liskovich, Diana and
                   Foss, Didem and Wang, Dingkang and Le, Duc and Holland,
                   Dustin and Dowling, Edward and Jamil, Eissa and Montgomery,
                   Elaine and Presani, Eleonora and Hahn, Emily and Wood, Emily
                   and Le, Eric-Tuan and Brinkman, Erik and Arcaute, Esteban and
                   Dunbar, Evan and Smothers, Evan and Sun, Fei and Kreuk, Felix
                   and Tian, Feng and Kokkinos, Filippos and Ozgenel, Firat and
                   Caggioni, Francesco and Kanayet, Frank and Seide, Frank and
                   Florez, Gabriela Medina and Schwarz, Gabriella and Badeer,
                   Gada and Swee, Georgia and Halpern, Gil and Herman, Grant and
                   Sizov, Grigory and {Guangyi} and {Zhang} and
                   Lakshminarayanan, Guna and Inan, Hakan and Shojanazeri, Hamid
                   and Zou, Han and Wang, Hannah and Zha, Hanwen and Habeeb,
                   Haroun and Rudolph, Harrison and Suk, Helen and Aspegren,
                   Henry and Goldman, Hunter and Zhan, Hongyuan and Damlaj,
                   Ibrahim and Molybog, Igor and Tufanov, Igor and Leontiadis,
                   Ilias and Veliche, Irina-Elena and Gat, Itai and Weissman,
                   Jake and Geboski, James and Kohli, James and Lam, Janice and
                   Asher, Japhet and Gaya, Jean-Baptiste and Marcus, Jeff and
                   Tang, Jeff and Chan, Jennifer and Zhen, Jenny and
                   Reizenstein, Jeremy and Teboul, Jeremy and Zhong, Jessica and
                   Jin, Jian and Yang, Jingyi and Cummings, Joe and Carvill, Jon
                   and Shepard, Jon and McPhie, Jonathan and Torres, Jonathan
                   and Ginsburg, Josh and Wang, Junjie and Wu, Kai and U, Kam
                   Hou and Saxena, Karan and Khandelwal, Kartikay and Zand,
                   Katayoun and Matosich, Kathy and Veeraraghavan, Kaushik and
                   Michelena, Kelly and Li, Keqian and Jagadeesh, Kiran and
                   Huang, Kun and Chawla, Kunal and Huang, Kyle and Chen, Lailin
                   and Garg, Lakshya and A, Lavender and Silva, Leandro and
                   Bell, Lee and Zhang, Lei and Guo, Liangpeng and Yu, Licheng
                   and Moshkovich, Liron and Wehrstedt, Luca and Khabsa, Madian
                   and Avalani, Manav and Bhatt, Manish and Mankus, Martynas and
                   Hasson, Matan and Lennie, Matthew and Reso, Matthias and
                   Groshev, Maxim and Naumov, Maxim and Lathi, Maya and
                   Keneally, Meghan and Liu, Miao and Seltzer, Michael L and
                   Valko, Michal and Restrepo, Michelle and Patel, Mihir and
                   Vyatskov, Mik and Samvelyan, Mikayel and Clark, Mike and
                   Macey, Mike and Wang, Mike and Hermoso, Miquel Jubert and
                   Metanat, Mo and Rastegari, Mohammad and Bansal, Munish and
                   Santhanam, Nandhini and Parks, Natascha and White, Natasha
                   and Bawa, Navyata and Singhal, Nayan and Egebo, Nick and
                   Usunier, Nicolas and Mehta, Nikhil and Laptev, Nikolay
                   Pavlovich and Dong, Ning and Cheng, Norman and Chernoguz,
                   Oleg and Hart, Olivia and Salpekar, Omkar and Kalinli, Ozlem
                   and Kent, Parkin and Parekh, Parth and Saab, Paul and Balaji,
                   Pavan and Rittner, Pedro and Bontrager, Philip and Roux,
                   Pierre and Dollar, Piotr and Zvyagina, Polina and
                   Ratanchandani, Prashant and Yuvraj, Pritish and Liang, Qian
                   and Alao, Rachad and Rodriguez, Rachel and Ayub, Rafi and
                   Murthy, Raghotham and Nayani, Raghu and Mitra, Rahul and
                   Parthasarathy, Rangaprabhu and Li, Raymond and Hogan,
                   Rebekkah and Battey, Robin and Wang, Rocky and Howes, Russ
                   and Rinott, Ruty and Mehta, Sachin and Siby, Sachin and
                   Bondu, Sai Jayesh and Datta, Samyak and Chugh, Sara and Hunt,
                   Sara and Dhillon, Sargun and Sidorov, Sasha and Pan, Satadru
                   and Mahajan, Saurabh and Verma, Saurabh and Yamamoto, Seiji
                   and Ramaswamy, Sharadh and Lindsay, Shaun and Lindsay, Shaun
                   and Feng, Sheng and Lin, Shenghao and Zha, Shengxin Cindy and
                   Patil, Shishir and Shankar, Shiva and Zhang, Shuqiang and
                   Zhang, Shuqiang and Wang, Sinong and Agarwal, Sneha and
                   Sajuyigbe, Soji and Chintala, Soumith and Max, Stephanie and
                   Chen, Stephen and Kehoe, Steve and Satterfield, Steve and
                   Govindaprasad, Sudarshan and Gupta, Sumit and Deng, Summer
                   and Cho, Sungmin and Virk, Sunny and Subramanian, Suraj and
                   Choudhury, Sy and Goldman, Sydney and Remez, Tal and Glaser,
                   Tamar and Best, Tamara and Koehler, Thilo and Robinson,
                   Thomas and Li, Tianhe and Zhang, Tianjun and Matthews, Tim
                   and Chou, Timothy and Shaked, Tzook and Vontimitta, Varun and
                   Ajayi, Victoria and Montanez, Victoria and Mohan, Vijai and
                   Kumar, Vinay Satish and Mangla, Vishal and Ionescu, Vlad and
                   Poenaru, Vlad and Mihailescu, Vlad Tiberiu and Ivanov,
                   Vladimir and Li, Wei and Wang, Wenchen and Jiang, Wenwen and
                   Bouaziz, Wes and Constable, Will and Tang, Xiaocheng and Wu,
                   Xiaojian and Wang, Xiaolan and Wu, Xilun and Gao, Xinbo and
                   Kleinman, Yaniv and Chen, Yanjun and Hu, Ye and Jia, Ye and
                   Qi, Ye and Li, Yenda and Zhang, Yilin and Zhang, Ying and
                   Adi, Yossi and Nam, Youngjin and {Yu} and {Wang} and Zhao, Yu
                   and Hao, Yuchen and Qian, Yundi and Li, Yunlu and He, Yuzi
                   and Rait, Zach and DeVito, Zachary and Rosnbrick, Zef and
                   Wen, Zhaoduo and Yang, Zhenyu and Zhao, Zhiwei and Ma, Zhiyu",
  journal       = "arXiv [cs.AI]",
  month         =  "31~" # jul,
  year          =  2024,
  archivePrefix = "arXiv",
  primaryClass  = "cs.AI",
  eprint        = "2407.21783"
}

@ARTICLE{lipshutz2016predicting,
  title     = "Predicting mortality in the intensive care unit: a comparison of
               the University Health Consortium expected probability of
               mortality and the Mortality Prediction Model {III}",
  author    = "Lipshutz, Angela K M and Feiner, John R and Grimes, Barbara and
               Gropper, Michael A",
  journal   = "J. Intensive Care",
  publisher = "Springer Nature",
  volume    =  4,
  number    =  1,
  pages     =  35,
  month     =  "23~" # may,
  year      =  2016,
  doi       = "10.1186/s40560-016-0158-z",
  pmc       = "PMC4876564",
  pmid      =  27217959,
  issn      = "2052-0492"
}

@ARTICLE{leGall1993new,
  title     = "A new Simplified Acute Physiology Score ({SAPS} {II}) based on a
               European/North American multicenter study",
  author    = "Le Gall, J R",
  journal   = "JAMA",
  publisher = "American Medical Association (AMA)",
  volume    =  270,
  number    =  24,
  pages     = "2957--2963",
  month     =  "22~" # dec,
  year      =  1993,
  doi       = "10.1001/jama.270.24.2957",
  pmid      =  8254858,
  issn      = "0098-7484,1538-3598"
}

@article{rosenblatt1956,
  title={Remarks on Some Nonparametric Estimates of a Density Function},
  author={Rosenblatt, Murray},
  journal={Annals of Mathematical Statistics},
  volume={27},
  pages={832--837},
  year={1956}
}

@article{parzen1962,
  title={On Estimation of a Probability Density Function and Mode},
  author={Parzen, Emanuel},
  journal={Annals of Mathematical Statistics},
  volume={33},
  pages={1065--1076},
  year={1962}
}

@article{GBDFracture2019,
  author  = {{GBD 2019 Fracture Collaborators}},
  title   = {Global, regional, and national burden of bone fractures in 204 countries and territories, 1990--2019: a systematic analysis from the Global Burden of Disease Study 2019},
  journal = {The Lancet Healthy Longevity},
  year    = {2021},
  volume  = {2},
  number  = {9},
  pages   = {e580--e592},
  doi     = {10.1016/S2666-7568(21)00172-0}
}

@article{Kanis2008FRAX,
  author  = {Kanis, J.A. and Johnell, O. and Od{\'e}n, A. and Johansson, H. and McCloskey, E.},
  title   = {FRAX and the assessment of fracture probability in men and women from the UK},
  journal = {Osteoporosis International},
  year    = {2008},
  volume  = {19},
  number  = {4},
  pages   = {385--397},
  doi     = {10.1007/s00198-007-0543-5}
}

@article{Kanis2009FRAXBone,
  author  = {Kanis, J.A. and Od{\'e}n, A. and Johansson, H. and Borgstr{\"o}m, F. and Str{\"o}m, O. and McCloskey, E.},
  title   = {FRAX and its applications to clinical practice},
  journal = {Bone},
  year    = {2009},
  volume  = {44},
  number  = {5},
  pages   = {734--743},
  doi     = {10.1016/j.bone.2009.01.373}
}

@misc{WHOFragility2024,
  author       = {{World Health Organization}},
  title        = {Fragility fractures},
  year         = {2024},
  note         = {Fact sheet, 25 September 2024},
  url          = {https://www.who.int/news-room/fact-sheets/detail/fragility-fractures}
}

@misc{USPSTF_OsteoporosisScreening_2025,
  author = {{U.S. Preventive Services Task Force}},
  title  = {Osteoporosis to Prevent Fractures: Screening},
  year   = {2025},
  note   = {Recommendation statement (Jan 14, 2025)},
  url    = {https://uspreventiveservicestaskforce.org/uspstf/recommendation/osteoporosis-screening}
}

@article{NicholsonJAMA2025,
  author  = {Nicholson, W. K. and others},
  title   = {Screening for Osteoporosis to Prevent Fractures: US Preventive Services Task Force Recommendation Statement},
  journal = {JAMA},
  year    = {2025}
}

@article{ching2018cox,
  title={Cox-nnet: an artificial neural network method for prognosis prediction of high-throughput omics data},
  author={Ching, Travers and Zhu, Xun and Garmire, Lana X},
  journal={PLoS computational biology},
  volume={14},
  number={4},
  pages={e1006076},
  year={2018},
  publisher={Public Library of Science San Francisco, CA USA}
}

@inproceedings{vaidyanathan2025survival,
  title={Survival Analysis for Cancers of the Brain, CNS and Bone using Retrieval Augmented Generation on the SEER Database},
  author={Vaidyanathan, Jyothi and Gupta, Shourya and Lee, Justin and Prabhu, Srikanth and Sengupta, Saptarshi},
  booktitle={Proceedings of the AAAI Symposium Series},
  volume={5},
  number={1},
  pages={31--36},
  year={2025}
}

@inproceedings{wen2024llm,
  title={LLM-Enhanced Survival Model for Electric Device Lifespan Estimation},
  author={Wen, Bao and Wen, Aihui and Fang, Wentian and Li, Jining},
  booktitle={2024 IEEE Smart World Congress (SWC)},
  pages={2547--2552},
  year={2024},
  organization={IEEE}
}

@article{luck2018learning,
  title={Learning to rank for censored survival data},
  author={Luck, Margaux and Sylvain, Tristan and Cohen, Joseph Paul and Cardinal, Heloise and Lodi, Andrea and Bengio, Yoshua},
  journal={arXiv preprint arXiv:1806.01984},
  year={2018}
}

@article{jing2019deep,
  title={A deep survival analysis method based on ranking},
  author={Jing, Bingzhong and Zhang, Tao and Wang, Zixian and Jin, Ying and Liu, Kuiyuan and Qiu, Wenze and Ke, Liangru and Sun, Ying and He, Caisheng and Hou, Dan and others},
  journal={Artificial intelligence in medicine},
  volume={98},
  pages={1--9},
  year={2019},
  publisher={Elsevier}
}

@article{shahid2026leveraging,
  title={Leveraging Large Language Models and Survival Analysis for Early Prediction of Chemotherapy Outcomes},
  author={Shahid, Muhammad Faisal and Afzal, Asad and Faiz, Abdullah and Siddiqui, Muhammad and Shehzad, Arbaz Khan and Aftab, Fatima and Shahid, Muhammad Usamah and Farooq, Muddassar},
  journal={arXiv preprint arXiv:2603.11594},
  year={2026}
}

@inproceedings{jeanselme2024review,
  title={Review of language models for survival analysis},
  author={Jeanselme, Vincent and Agarwal, Nikita and Wang, Chen},
  booktitle={AAAI 2024 Spring Symposium on Clinical Foundation Models},
  year={2024}
}

@inproceedings{lee2018deephit,
  title={Deephit: A deep learning approach to survival analysis with competing risks},
  author={Lee, Changhee and Zame, William and Yoon, Jinsung and Van Der Schaar, Mihaela},
  booktitle={Proceedings of the AAAI conference on artificial intelligence},
  volume={32},
  number={1},
  year={2018}
}

@article{katzman2018deepsurv,
  title={DeepSurv: personalized treatment recommender system using a Cox proportional hazards deep neural network},
  author={Katzman, Jared L and Shaham, Uri and Cloninger, Alexander and Bates, Jonathan and Jiang, Tingting and Kluger, Yuval},
  journal={BMC medical research methodology},
  volume={18},
  number={1},
  pages={24},
  year={2018},
  publisher={Springer}
}

\newpage
\appendix
\setcounter{subsection}{0}
\setcounter{table}{0}
\setcounter{figure}{0}
\renewcommand{\tablename}{eTable}
\renewcommand{\figurename}{eFigure}


\part*{Supplementary materials}
\label{sec:suppl}

\begin{table}[H]
\centering
\caption{Summary of previous literature}
\label{tab:survival_analysis_literature}
\rowcolors{2}{white}{gray!20}
\small
\begin{tabularx}{\textwidth}{
l
>{\raggedright\arraybackslash}p{7em}
>{\raggedright\arraybackslash}X
}
\toprule
\textbf{Author} & \textbf{Model Type} & \textbf{Description of how they tackle survival analysis} \\
\midrule
\citet{esteban2024predictive} & Deep Learning, LLM & Added dedicated neural networks on top of an LLM and experimented with ordinal regression loss and discordant pair loss. \\
\citet{jeanselme2024review} & LLM (Review) & Reviewed existing literature, observing that most studies utilize LLMs as feature extractors rather than direct predictive models, or require substantial architectural modifications. \\
\citet{jing2019deep} & Deep Learning & Tackling the survival analysis problem as a ranking problem in the deep learning Era. \\
\citet{luck2018learning} & Deep Learning & Tackling the survival analysis problem as a ranking problem in the deep learning Era. \\
\citet{shahid2026leveraging} & LLM & Used an LLM as a feature extractor to obtain variables from clinical notes. \\
\citet{steinberg2023motor} & Deep Learning, LLM & Attached a piecewise exponential artificial neural network on top of a foundation model dedicated to predicting the hazard rate. \\
\citet{vaidyanathan2025survival} & LLM & Simplified the survival analysis problem into a five-year classification task and did not address censoring. \\
\citet{Van-Belle2011-ab} & Machine Learning & Explored treating survival analysis as a ranking problem using support vector methods. \\
\citet{wen2024llm} & LLM & Used an LLM to extract features for a downstream survival model. \\
\bottomrule
\end{tabularx}
\end{table}

\newpage

\begin{table}[ht]
\caption{Performance comparison for ICU mortality prediction on the MIMIC-ICU cohort.}
\centering
\footnotesize
\begin{tabular}{lllll}
\toprule
 & Overall & 1 day & 2 day & 7 day \\ 
& (n=1,984) & (n=1,955) & (n=1,885) & (n=1,239) \\ 
\midrule
\rowcolor{gray!20}\textbf{C-Index} &&&& \\
SAPS II & 0.746 (0.719, 0.774) & 0.771 (0.724, 0.820) & 0.784 (0.742, 0.824) & 0.726 (0.695, 0.756) \\ 
Cox & 0.753 (0.726, 0.779) & 0.784 (0.729, 0.834) & 0.793 (0.747, 0.835) & 0.744 (0.713, 0.775) \\ 
Cox-nnet & 0.770 (0.744, 0.796) & 0.820 (0.774, 0.863) & 0.819 (0.778, 0.856) & 0.760 (0.730, 0.787) \\ 
Deephit & 0.730 (0.701, 0.756) & 0.743 (0.681, 0.800) & 0.755 (0.706, 0.802) & 0.733 (0.701, 0.767) \\
Deepsurv & 0.766 (0.739, 0.791) & 0.816 (0.766, 0.863) & 0.825 (0.786, 0.862) & 0.756 (0.727, 0.784) \\
\Ourmodel(Qwen) & 0.784 (0.761, 0.807) & 0.832 (0.787, 0.872) & 0.841 (0.806, 0.872) & 0.772 (0.745, 0.797) \\ 
\Ourmodel(Llama) & 0.776 (0.752, 0.800) & 0.824 (0.779, 0.865) & 0.828 (0.791, 0.862) & 0.765 (0.737, 0.791) \\ 
Logistic regression + ranking & 0.749 (0.716, 0.776) & 0.770 (0.704, 0.830) & 
0.782 (0.732, 0.828) & 0.738 (0.703, 0.770) \\

\midrule
\rowcolor{gray!20}\textbf{AUC} &&&& \\ 
SAPS II & 0.792 (0.768, 0.814) & 0.779 (0.730, 0.828) & 0.795 (0.752, 0.836) & 0.753 (0.720, 0.785) \\ 
Cox model & 0.783 (0.759, 0.807) & 0.791 (0.734, 0.842) & 0.804 (0.757, 0.847) & 0.771 (0.739, 0.805) \\ 
Cox-nnet & 0.798 (0.774, 0.820) & 0.826 (0.779, 0.869) & 0.829 (0.787, 0.867) & 0.785 (0.754, 0.815) \\ 
Deephit & 0.743 (0.716, 0.767) & 0.749 (0.684, 0.807) & 0.763 (0.713, 0.812) & 0.760 (0.727, 0.795) \\
Deepsurv & 0.795 (0.773, 0.818) & 0.823 (0.772, 0.870) & 0.835 (0.795,0.873)&0.780 (0.750, 0.810)\\
\Ourmodel(Qwen) & 0.809 (0.787, 0.830) & 0.839 (0.793, 0.880) & 0.853 (0.817, 0.886) & 0.797 (0.767, 0.825) \\ 
\Ourmodel(Llama) & 0.805 (0.783, 0.826) & 0.831 (0.786, 0.872) & 0.839 (0.800, 0.875) & 0.792 (0.762, 0.820) \\ 
Logistic regression + ranking & 0.785 (0.757, 0.810) & 0.777 (0.709, 0.836) & 0.793 (0.740, 0.840) & 0.764 (0.727, 0.800) \\

\bottomrule
\end{tabular}
\label{tab_ICU_results}
\end{table}

\newpage

\begin{table}[ht]
\caption{Performance comparison for fracture risk assessment on the NYP/WCM fracture cohort.}
\centering
\footnotesize
\begin{tabular}{lllll}
\toprule
 & Overall & 2y & 3y & 5y \\ 
& (n=2,235) & (n=1,738) & (n=1,244) & (n=322) \\ 
\midrule
\rowcolor{gray!20}\textbf{C-Index} &&&& \\ 
FRAX & 0.660 (0.614, 0.701) & 0.659 (0.613, 0.703) & 0.661 (0.618, 0.701) & 0.622 (0.583, 0.663) \\ 
Cox model& 0.737 (0.694, 0.782) & 0.762 (0.709, 0.813) & 0.742 (0.694, 0.785) & 0.697 (0.651, 0.739) \\ 
Cox-nnet & 0.684 (0.632, 0.728) & 0.718 (0.659, 0.774) & 0.689 (0.642, 0.737) & 0.675 (0.630, 0.718) \\ 
Deephit & 0.729 (0.681, 0.773) & 0.763 (0.705, 0.815) & 0.743 (0.696, 0.789) & 0.723 (0.682, 0.761) \\
DeepSurv & 0.730 (0.682, 0.772) & 0.766 (0.711, 0.816) & 0.742 (0.697, 0.785) & 0.708 (0.666, 0.747) \\
\Ourmodel(Qwen) & 0.741 (0.699, 0.784) & 0.772 (0.723, 0.818) & 0.745 (0.699, 0.786) & 0.705 (0.662, 0.746) \\ 
\Ourmodel(Llama) & 0.742 (0.699, 0.783) & 0.774 (0.723, 0.818) & 0.744 (0.696, 0.784) & 0.701 (0.657, 0.740) \\ 
Logistic regression + ranking & 0.738 (0.691, 0.788) & 0.766 (0.709, 0.812) & 0.723 (0.677, 0.774) & 0.699 (0.628, 0.770) \\
\midrule
\rowcolor{gray!20}\textbf{AUC} &&&& \\ 
FRAX & 0.664 (0.621, 0.704) & 0.665 (0.616, 0.710) & 0.672 (0.626, 0.715) & 0.681 (0.623, 0.735) \\ 
Cox model& 0.717 (0.671, 0.761) & 0.767 (0.713, 0.820) & 0.751 (0.700, 0.796) & 0.735 (0.676, 0.790) \\ 
Cox-nnet & 0.647 (0.600, 0.692) & 0.721 (0.661, 0.778) & 0.693 (0.643, 0.742) & 0.696 (0.635, 0.755) \\ 
Deephit & 0.693 (0.647, 0.739) & 0.767 (0.708, 0.821) & 0.749 (0.701, 0.798) & 0.756 (0.703, 0.804) \\
DeepSurv & 0.697 (0.649, 0.740) & 0.771 (0.714, 0.821) & 0.749 (0.702, 0.794) & 0.742 (0.688, 0.792) \\
\Ourmodel(Qwen) & 0.720 (0.678, 0.761) & 0.777 (0.727, 0.825) & 0.753 (0.704, 0.796) & 0.741 (0.684, 0.795) \\ 
\Ourmodel(Llama) & 0.721 (0.676, 0.761) & 0.779 (0.727, 0.824) & 0.753 (0.703, 0.795) & 0.742 (0.685, 0.796) \\ 
Logistic regression + ranking & 0.714 (0.666, 0.764) & 0.769 (0.710, 0.821) & 0.730 (0.676, 0.779) & 0.821 (0.717, 0.905) \\
\bottomrule
\end{tabular}
\label{tab_Fracture_results}
\end{table}

\newpage

\begin{table}[H]
\centering
\caption{Summary of covariates used for ICU mortality prediction on  MIMIC-IV ICU dataset.}
\rowcolors{2}{white}{gray!20}
\label{covariate_icu}
\begin{tabular}{ll}
\toprule
\textbf{Covariate} & \textbf{Description} \\
\midrule
age\_score & Points assigned based on patient age. \\
hr\_score & Heart rate abnormality score. \\
sysbp\_score & Systolic blood pressure abnormality score. \\
temp\_score & Body temperature abnormality score. \\
PaO2FiO2\_score & Oxygenation score based on the PaO\textsubscript{2}/FiO\textsubscript{2} ratio. \\
uo\_score & Urine output score reflecting renal perfusion. \\
bun\_score & Blood urea nitrogen abnormality score (renal function). \\
wbc\_score & White blood cell count abnormality score (infection/sepsis). \\
potassium\_score & Serum potassium abnormality score. \\
sodium\_score & Serum sodium abnormality score. \\
bicarbonate\_score & Serum bicarbonate abnormality score (acid--base balance). \\
bilirubin\_score & Serum bilirubin level score (liver dysfunction). \\
gcs\_score & Glasgow Coma Scale score (neurological status). \\
comorbidity\_score & Score derived from chronic health conditions. \\
admissiontype\_score & Points assigned based on admission type. \\
\bottomrule
\end{tabular}
\end{table}

\newpage

\begin{table}[H]
\centering
\caption{Summary of covariates used for fracture risk assessment on  NPY/WCM-fracture dataset}
\label{covariate_fracture}
\rowcolors{2}{white}{gray!20}
\begin{tabularx}{\textwidth}{lX}
\toprule
\textbf{Covariate} & \textbf{Description} \\
\midrule
Age & Age at baseline.\\
Sex & Participant's sex \\
Previous\_fracture& Whether the patient has had a prior fracture.\\
T-score (NECK AVG) category & T-score from the average bone density of the femoral neck, categorized to Normal (above -1), Osteopenia (-2.5~-1.0), and Osteoporosis (below -2.5). \\
Osteoporosis-related disease count & Count of the 22 diseases listed in Table \ref{tab:fracture}. \\
Primidone & An anticonvulsant medication used to treat epilepsy and essential tremor. \\
Dexamethasone & A potent synthetic corticosteroid (steroid) medication used to treat a wide range of inflammatory, autoimmune, and hormonal conditions. \\
Prednisone & A potent synthetic corticosteroid used to treat inflammatory and autoimmune conditions, ranging from asthma to cancer.\\
Degarelix & a GnRH antagonist injection for advanced prostate cancer. \\
Atazanavir & An antiretroviral protease inhibitor for HIV.\\
\bottomrule
\end{tabularx}
\end{table}

\newpage 
\begin{table}[H]
\centering
\caption{Hyperparameters and tuning protocol for baselines}
\label{tab:baseline_hyperparams}
\rowcolors{2}{white}{gray!20}
\setlength{\extrarowheight}{2pt}
\begin{tabularx}{\textwidth}{
    >{\raggedright\arraybackslash}p{0.15\textwidth}
    >{\raggedright\arraybackslash}X
    >{\raggedright\arraybackslash}p{0.28\textwidth}}
\toprule
\textbf{Method} & \textbf{Hyperparameters} & \textbf{Selection Protocol} \\
\midrule

CoxNet &
\texttt{alphas} $= [0.1, 0.01, 0.001, 0.0001]$; \texttt{l1\_ratio} $= 0.5$;
\texttt{fit\_baseline\_model} $=$ \texttt{True} &
Elastic-net regularization path specified before final testing. \\

Cox-nnet &
\texttt{L2\_reg} $= 0.001$; hidden layers $[64, 32]$; dropout $0.2$;
learning rate $0.001$; optimizer/search method \texttt{nesterov};
\texttt{eval\_step} $= 50$; \texttt{max\_iter} $= 10000$;
early-stopping patience $1000$; random seed $42$ &
Training controlled by fixed hyperparameters and early stopping. \\

DeepSurv &
hidden layers $[64, 32]$; batch normalization \texttt{True}; dropout $0.2$;
batch size $256$; learning rate \texttt{1e-3}; maximum epochs $300$;
random seed $42$ &
Trained with an internal 80/20 training--validation split and early stopping. \\

DeepHit &
number of duration bins $20$; hidden layers $[32, 16]$;
batch normalization \texttt{True}; dropout $0.2$; batch size $128$;
learning rate \texttt{5e-4}; maximum epochs $300$; \texttt{alpha} $= 0.5$;
\texttt{sigma} $= 0.2$; random seed $42$ &
Trained with an internal 80/20 training--validation split and early stopping. \\

\bottomrule
\end{tabularx}
\end{table}

\newpage

\begin{table}[H]
\centering
\caption{Average coefficient of variation across different anchor point selections}
\label{tab:avg_cv}
\begin{tabular}{llll}
\toprule
\textbf{Average coefficient of variation} & \textbf{Random} & \textbf{Positive only} & \textbf{Random + Positive only} \\
\midrule
MIMIC-ICU        & 0.1410 & 0.1410 & 0.1414 \\
NYP/WCM Fracture & 0.1412 & 0.1408 & 0.1413 \\
\bottomrule
\end{tabular}
\end{table}

\newpage

\section{Anchor selection minimally influences risk predictions}
\label{sec:anchor selection}

\begin{figure}[h]
    \centering
    \includegraphics[width=.7\linewidth]{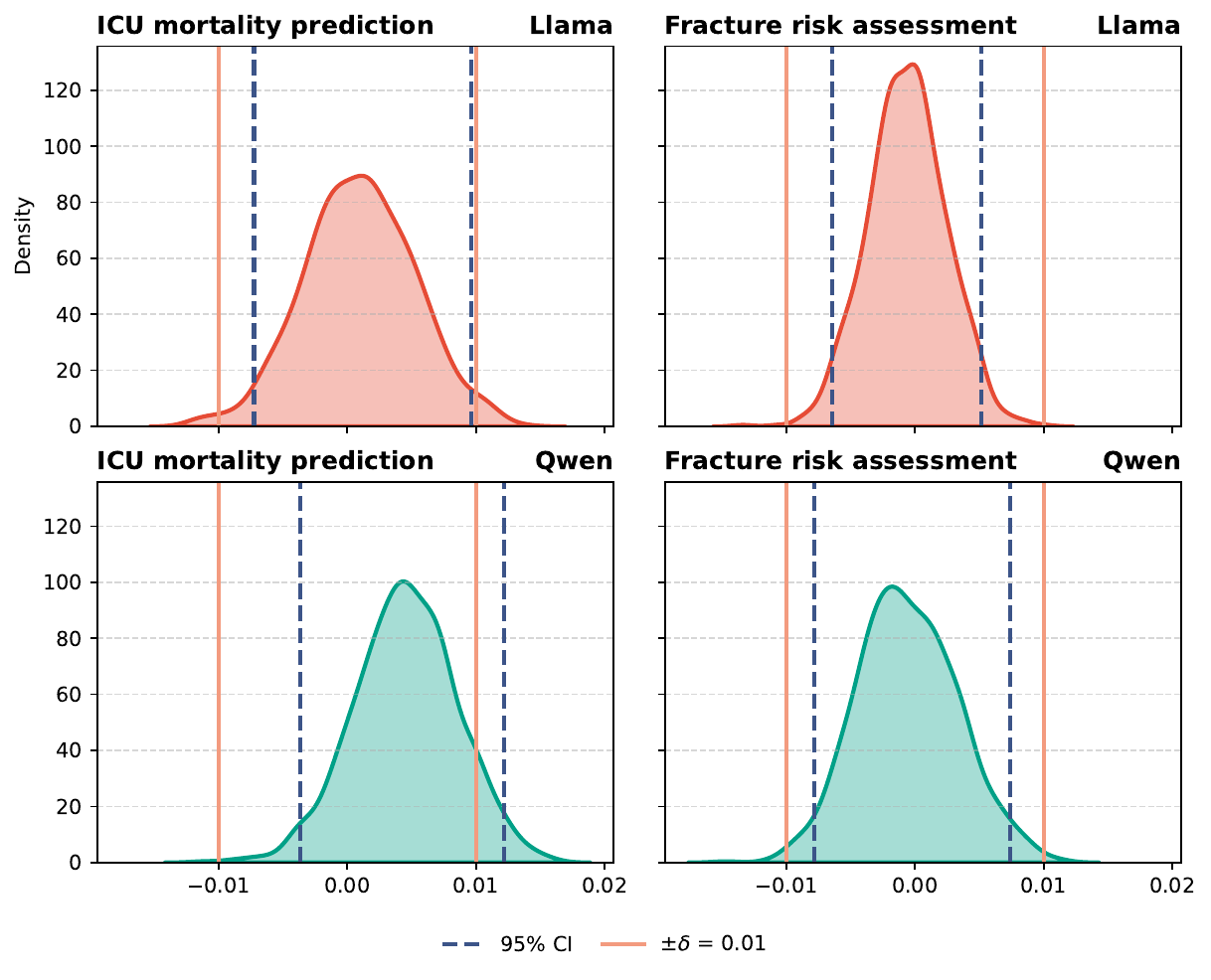}
    
    \caption{\textbf{Comparison of anchor selection strategies for risk estimation.} Paired differences in C-index between random anchor selection and event-only anchors (random minus event-only) using LLaMA and Qwen backbones. 
Kernel density estimation (KDE) curves depict bootstrap distributions of paired differences, with shaded regions indicating 95\% confidence intervals. Dashed lines denote equivalence bounds ($\pm\delta = 0.01$).
}
    \label{fig:KDE_selection}
\end{figure}

We assessed whether anchor-point selection influences risk estimation (eFigure~\ref{fig:KDE_selection}). We compared two strategies: (1) random sampling of anchors from the full training set (``random''; primary analysis) and (2) selection of anchors exclusively from event cases (``event-only''). Paired differences in C-index between the two strategies were visualized using kernel density estimation, together with bootstrap confidence intervals and an equivalence bound of $\delta = 0.01$.

Across both tasks, differences were minimal for both Llama- and Qwen-based \Ourmodel. For Llama-based \Ourmodel, the mean differences were 0.0010 (-0.0072, 0.0096) for ICU mortality prediction and -0.0006 (-0.0064, 0.0051) for fracture risk assessment. For Qwen-based \Ourmodel, the corresponding mean differences were 0.0045 (-0.0037, 0.0122) and 0.0031 (-0.0143, 0.0218). Most estimates lie within the equivalence bound, indicating similar performance across anchor-selection strategies. Even when differences slightly exceeded $\delta$, the absolute deviations were negligible, and the confidence intervals remained narrow, supporting the conclusion that anchor selection has little influence on risk estimation.

\vspace{1em}

\newpage

\section{\Ourmodel is robust to subject order in LLM ranking}
\label{sec:order}

Order inconsistency is a known limitation of LLMs as rankers \cite{Qin2024-eo, Zeng2024-ye}, referring to cases in which reversing the order of subjects in the prompt (for example, ``rank A, B'' versus ``rank B, A'') leads to different outputs. To evaluate the impact of this issue, we compared two prompting strategies: (1) randomly shuffling the target and anchor subjects (``shuffle''; primary analysis) and (2) fixing the anchor in the first position (``fix\_a''). Paired differences in C-index between the two strategies were visualized using kernel density estimation (KDE) \cite{rosenblatt1956,parzen1962} with bootstrap resampling, together with 95\% confidence intervals and an equivalence bound of $\delta = 0.01$ to assess practical significance.

\begin{figure}[h]
    \centering
    \includegraphics[width=.7\linewidth]{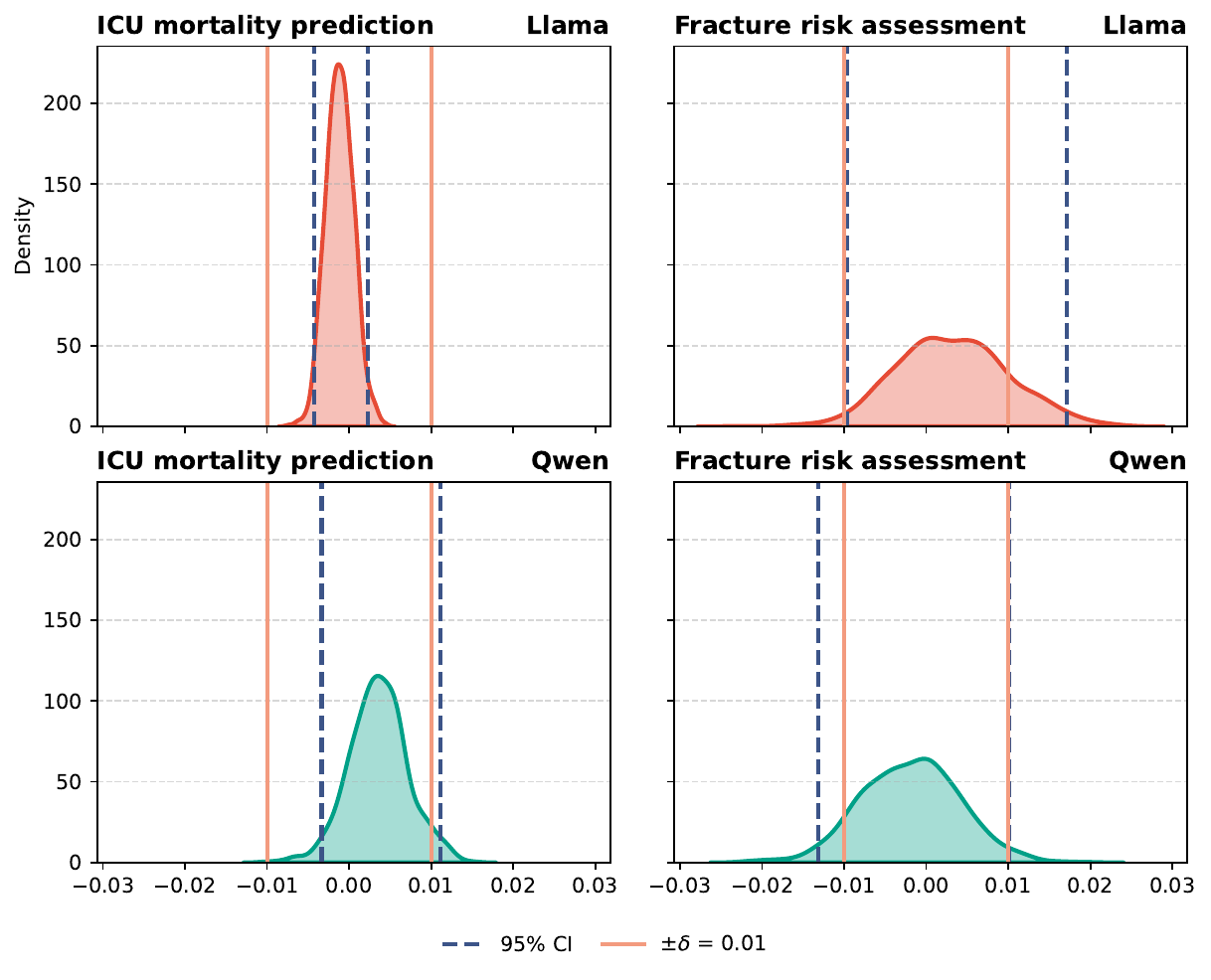}
    \caption{Evaluation of order inconsistency in pairwise comparisons. Paired differences in C-index between shuffled and fixed-anchor ordering (shuffle minus fix\_a) using LLaMA and Qwen backbones. 
Kernel density estimation (KDE) curves depict bootstrap distributions of paired differences, with shaded regions indicating 95\% confidence intervals. Dashed lines denote equivalence bounds ($\pm\delta = 0.01$).
}
    \label{fig:KDE_order}
\end{figure}

As shown in eFigure~\ref{fig:KDE_order}, differences were minimal across tasks. For Llama-based \Ourmodel, the mean paired differences were -0.0011 (-0.0043, 0.0023) for ICU mortality prediction and 0.0033 [-0.0097, 0.0171] for fracture risk assessment. Relative to the equivalence bound, the difference for ICU mortality fell well within the prespecified threshold, indicating negligible effects of subject ordering. For fracture risk assessment, the estimate slightly exceeded $\delta$, but the absolute deviation remained small. Bootstrap confidence intervals were narrow and centered near zero, supporting comparable performance between the two strategies. Similar patterns were observed for Qwen-based \Ourmodel, with mean paired differences of 0.0036 (-0.0034, 0.0112) and -0.0004 (-0.0192, 0.0106). Overall, these findings suggest that \Ourmodel is robust to subject order, with minimal influence on risk estimation performance.

\vspace{1em}







\end{document}